# Graphemic Normalization of the Perso-Arabic Script

Raiomond Doctor,* Alexander Gutkin, Cibu Johny, Brian Roark, and Richard Sproat
Google Research: India, United Kingdom, United States, and Japan
{raiomond,agutkin,cibu,roark,rws}@google.com

**Abstract**

Since its original appearance in 1991, the Perso-Arabic script representation in Unicode has grown from 169 to over 440 atomic isolated characters spread over several code pages representing standard letters, various diacritics and punctuation for the original Arabic and numerous other regional orthographic traditions (Unicode Consortium, 2021). This paper documents the challenges that Perso-Arabic presents beyond the best-documented languages, such as Arabic and Persian, building on earlier work by the expert community (ICANN, 2011, 2015). We particularly focus on the situation in natural language processing (NLP), which is affected by multiple, often neglected, issues such as the use of visually ambiguous yet canonically nonequivalent letters and the mixing of letters from different orthographies. Among the contributing conflating factors are the lack of input methods, the instability of modern orthographies (e.g., Aazim et al., 2009; Iyengar, 2018), insufficient literacy, and loss or lack of orthographic tradition (Jahani and Korn, 2013; Liljegren, 2018). We evaluate the effects of script normalization on eight languages from diverse language families in the Perso-Arabic script diaspora on machine translation and statistical language modeling tasks. Our results indicate statistically significant improvements in performance in most conditions for all the languages considered when normalization is applied. We argue that better understanding and representation of Perso-Arabic script variation within regional orthographic traditions, where those are present, is crucial for further progress of modern computational NLP techniques (Ponti et al., 2019; Conneau et al., 2020; Muller et al., 2021) especially for languages with a paucity of resources.

## 1 Introduction

The Modern Perso-Arabic script derives from the fourth century North Arabic script, which in turn was adapted from the Nabatean Aramaic script to write the Arabic language (Gruendler, 1993; Bauer, 1996). Due to the spread of Islam throughout much of Africa, Asia, and parts of Europe, it has come, in its various adapted forms, to be one of the most widely used scripts in the modern world. Due to its reasonable flexibility in representing phonological structure, the script was adopted to write a large number of languages spanning diverse language families such as Afro-Asiatic, Indo-European, Niger-Congo, Turkic, and Sino-Tibetan, among others. Adaptations are found as far south as Southern Africa with Arabic having been used for Afrikaans (Kotzé, 2012) and Malagasy (Versteegh, 2001); as far east as East Asia for Chinese (Suutarinen, 2013) and Japanese (Naim, 1971; Kaye, 1996); and into Eastern Europe for writing languages of Muslim Slavs, such as Bosnian (Buljina, 2019). While many of these adaptations have not survived, the Arabic script and its derivatives are still used for scores of languages with a total population of speakers of over 600 million.[1] For some linguistic areas, such as the Dardic languages of Northern Pakistan, most of which were unwritten until very recently, the Perso-Arabic script is the *only* serious contender when developing a new writing system; see for example Torwali (Torwali, 2019),[2] and Palula (Liljegren, 2016).

While the original Semitic scripts were pure consonantal scripts (*abjads*), three letters—*alif* /ʔ/, *ya* /y/, and *waw* /w/—came to be used as *matres lectionis* to represent long vowels, and further diacritics were developed to (optionally) represent such features as short vowels and gemination (*shadda*), among others (Bauer, 1996).

*On contract from Optimum Solutions, Inc.
[1] https://en.wikipedia.org/wiki/List_of_writing_systems#List_of_writing_systems_by_adoption
[2] https://www.blog.google/around-the-globe/google-asia/torwali-language-and-its-new-android-keyboard/

The original North Arabic script was rather ambiguous, since Arabic had a larger consonant inventory than Aramaic, and some of the consonant letters had to do double duty – a problem exacerbated by the cursivization of the script. The resulting ambiguities were resolved by the use of various numbers of dots over or under the letters to disambiguate the various uses (Bauer, 1996; Kaplony, 2008), a system called *i'jām* (إِعْجَام). For example the inferior dot in ⟨ب⟩ /b/ distinguishes it from ⟨ن⟩ /n/ with a single dot on top, and then again from ⟨ت⟩ /t/ with two dots on top, and again still from ⟨ث⟩ /θ/ with three on top. Though the set of consonants to be disambiguated is of course limited in Arabic itself, the *i'jām*, once started, evolved into a productive way to produce new consonant symbols when the script was adapted to new languages. This has consequently allowed languages to have their "own" version of the Perso-Arabic script, where the only difference with the scripts used for a language's neighbors is in the use of distinctive *i'jām*-augmented consonants. This is true, for example, for adaptations of the script to the many Dardic languages, where each has one or two consonant symbols not found in the scripts of its neighbors.

As noted above, the Arabic script and its derivatives include diacritics that allow one to specify all vowels and other phonetic features such as gemination. However in the normal daily use of the script for Arabic and other languages, these are typically omitted. In Arabic this means that the script is still technically an *abjad*, since the written symbols mostly represent consonants. However to varying degrees, the derived scripts have departed from this, and some of them are full alphabets. Thus according to Kaye (1996), the Persian writing system is an *abjad* as are Urdu and Jawi, the old Malay Arabic-based writing system; however the Kurdish and Uyghur writing systems are alphabets. Parallel developments occurred with adaptations of the Hebrew *abjad* so that Yiddish orthography (Aronson, 1996) is an alphabet.

Historically each geographic region posed its own unique sociolinguistic challenges resulting in the emergence of different adaptation strategies and orthographic traditions across South Asia (Wink, 1991; Qutbuddin, 2007), Southeast Asia (Kratz, 2002; Ricci, 2011; Abdullah et al., 2020), and Africa (Mumin, 2014; Ngom and Kurfi, 2017), among other regions (Suutarinen, 2013; de Castilla, 2019). This diversity as well as the flexible nature of the script is reflected in a large and growing inventory of Perso-Arabic code points in the Unicode standard (Unicode Consortium, 2021) with accompanying ambiguities associated with representing the script using the digital medium that we briefly outline in §2. We then provide an overview of some of the regional orthographies for eight languages selected from diverse language families in §3. A significant amount of digital representation ambiguities manifest by these orthographies is resolved computationally using finite-state normalization methods for Perso-Arabic, described in §4, that we developed for this purpose.[3] We study the effects of normalization of real-world text using statistical and neural techniques, and present our findings in §5.1 and §5.2, respectively. The code and the results accompanying the experiments have been released.[4]

## 2 Perso-Arabic in the Digital Medium

As was mentioned previously, an important feature which has led to the adoption of the script by different cultures to use it to transcribe their language, is its very flexibility.[5] At its core the Arabic script comprises 18 basic shapes often referred to as *rasm* (رَسْم) or "drawing" (Daniels, 2013; Kurzon, 2013). These can be modified in various ways: apposing one to four dots (*i'jām*) placed above, below or inside a character (as shown in Figure 1); using modifier signs such as the subscript or superscript *small hamza*; placing diacritics or *tashkīl* (تَشكِيل) and in certain cases even adding a new shape based on the basic Arabic template. Thus for example Urdu, discussed in more detail in §3.1, substantially expanded the original Arabic writing system adapting it to its phonology by introducing additional *i'jām* characters, modifiers, and even creating new shapes such as the *bari yeh* ⟨ے⟩ or the *heh do chashmee* ⟨ھ⟩ for handling aspiration.

[3] https://github.com/google-research/nisaba

[4] https://github.com/google-research/google-research/tree/master/perso_arabic_norm

[5] Scripts are sets of characters used jointly in written representation while writing systems additionally consist of the rules and conventions used when employing a script for a particular language.

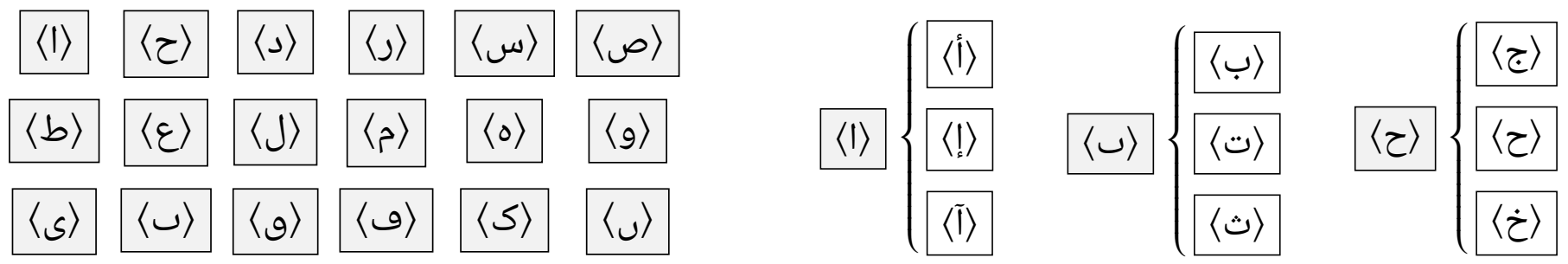

(a) (b)

Figure 1: Core *rasm* shapes, or *archigraphemes* according to Milo (2002), of Arabic shown in (a), and examples of Arabic letters derived with *i'jām* demonstrating their disambiguation function in (b), after (Nemeth, 2017).

| Display | Unicode Character Sequence $C_1$ | $C_2$ | $C_3$ | $C_4$ | $C_5$ | Transformation |
|---|---|---|---|---|---|---|
| رئيس | *reh* U+0631 | *yeh with hamza above* U+0626 | *yeh* U+064A | *seen* U+0633 | | |
| رئيس | *reh* U+0631 | *yeh* U+064A | *hamza above* U+0654 | *yeh* U+064A | *seen* U+0633 | Unicode NFC |
| رئيس | *reh* U+0631 | *alef maksura* U+0649 | *hamza above* U+0654 | *yeh* U+064A | *seen* U+0633 | *Visual* Normalization |
| رئيس | *reh* U+0631 | *yeh with hamza above* U+0626 | *farsi yeh* U+06CC | *seen* U+0633 | | *Visual* Normalization |
| رئيس | *reh* U+0631 | *farsi yeh* U+06CC | *hamza above* U+0654 | *yeh* U+064A | *seen* U+0633 | *Reading* Normalization |
| رئيس | *reh* U+0631 | *farsi yeh* U+06CC | *hamza above* U+0654 | *farsi yeh* U+06CC | *seen* U+0633 | *Reading* Normalization |

Table 1: Six different spellings of the Arabic word for "leader" (MSA: /ra.ʔiːs/) rendered in Naskh. For each row, the Unicode character differences with the Unicode string in the first row are highlighted. The last column indicates the type of transformation required to bring the relevant Unicode string to the canonical form displayed in the first row of the table.

Similar to Brahmic scripts, the Perso-Arabic script often provides more than one way to compose a character in the digital medium (Unicode Consortium, 2021). For example, the *alef with madda above* letter can be composed in two ways: as a single character ⟨آ⟩ (U+0622) or by adjoining *madda above* to *alef* (U+0627 followed by U+0653). This results in presentation ambiguity and the Unicode standard provides a certain number of canonical *normalization* forms, such as the Normalization Form C, or NFC, to handle such cases (Whistler, 2021). A normalization process is required to convert strings to such canonical forms. In contrast to Brahmic script normalization, where atomic forms are normalized to their decomposition, Perso-Arabic normalization in NFC involves combining decomposed elements into a single glyph. Thus the individual *alef* and *madda above* will be normalized to a single glyph. Our investigations have found many cases of this kind of *visual ambiguity* in the Perso-Arabic script beyond what is covered in NFC.

Some of these visual ambiguities are illustrated by a simple example in Table 1, where six alternate representations for the Arabic word for "leader" are shown rendered in Naskh style along with the corresponding Unicode sequences ranging from four to five characters. The spelling in the first row of the table is the correct spelling of the word in Modern Standard Arabic (MSA) orthography. The visual forms in the second, third and fourth rows are visually identical to the correct spelling, but are represented digitally as distinct sequences of characters. The second example, while ambiguous, is handled by the Unicode NFC normalization, which brings it to the canonical form provided by the first row by rewriting a (decomposed) sequence of *yeh* and *hamza above* as its canonical single-letter counterpart *yeh with hamza above*. The form in the third row is more problematic. It arises from a five-character sequence which has *alef maksura* instead of *yeh* and is also visually identical to words in rows one and two. However, unlike the spelling in the second row, the sequence *alef maksura* followed by *hamza above* does not have a canonical composed form under Unicode.[6] Hence, while this form is visually identical to the first and second forms, it is treated as a distinct word in the digital medium. Similarly, the example in the fourth row is visually identical to the preceding examples, but arises due to using *farsi yeh* instead of *yeh* which is illegal under MSA orthography. No standard transformation is provided by the Unicode to cope with cases like this example as they are orthography- and language-specific. We refer to the class of normalizations that result in something that is visually identical as *visual* normalization (Johny et al., 2021).

[6]See the discussion in (Pournader, 2010) on how this came about.

The second group of ambiguities is illustrated by the last two rows of Table 1. The fifth and sixth examples, while visually identical to each other, differ slightly from the reference in extra *i'jām* dots produced by combining *farsi yeh* with *hamza above* word-internally. Similar to the examples in rows three and four, no fallback normalization strategy is provided by the Unicode standard to handle such cases as it is not clear without prior context which orthography is intended. We refer to the class of normalizations that result in something that is not visually identical as *reading* normalization (Gutkin et al., 2022b).

As we mentioned above, unlike the canonical Unicode transformations, it is impossible to define most of visual and reading normalizations outside the orthographic context. Consider a sequence consisting of *waw* (U+0648) followed by *damma* (U+064F) whose visual form ⟨وُ⟩ is identical to letter *u* (U+06C7) used in Kazakh, Uzbek and Uyghur Perso-Arabic orthographies among some others (Aqtay, 2020; Haralambous, 2021). Normalization of *waw* and *damma* to its "canonical" form *u* should only be performed for these languages. We introduce the visual and reading script normalization framework more formally in §4.

One could argue that the ambiguities described above are not problematic and are the natural outcome of the specific properties of the script (e.g., its cursive form and the presence of positional variants), the vast number of orthographic adaptations and specifics of its implementation in digital medium. However, as we show in §5, the resolution of such ambiguities on a language-by-language basis positively impacts the quality of computational models of natural language. Furthermore, the visual ambiguities manifest by various Perso-Arabic writing systems represent a growing challenge to cybersecurity. From the standards' perspective this is being gradually addressed by the Internet Corporation for Assigned Names and Numbers (ICANN) by developing a robust set of standards for representing various Internet entities in Perso-Arabic script, such as domain names, with particular focus on visually confusable character variants (ICANN, 2011). In addition, security implications, such as development of mechanisms for protection against *phishing* and *spoofing* attacks, are actively studied by the relevant cybersecurity literature on Perso-Arabic (Hussain et al., 2016; Elsayed and Shosha, 2018; Ahmad and Erdodi, 2021).

A detailed analysis of the causes for the various types of Perso-Arabic script ambiguities described above are outside the scope of this work. It is however worthwhile to briefly mention some of them here. One of the causes is the relative complexity and several special properties of the Perso-Arabic script itself. The script has four key orthographic properties that are relevant here: (i) relative visual similarity of the *rasm* forms; (ii) *allography*, i.e., positional variants of letters (initial, medial, final and isolated); (iii) cursivity/ligaturing, and (iv) *non-linearity*, the extensive and sophisticated use of various types of *tashkīl* and *i'jām* (Yassin et al., 2020). The combination of all these properties was demonstrated to lead to relatively more involved visual processing of the script (compared to languages that use the Latin script) in the psycholinguistics and neuropsychology literature on reading for MSA (Eviatar and Ibrahim, 2014; Hermena and Reichle, 2020; Boudelaa et al., 2020), but also for the Uyghur Perso-Arabic orthography (Yakup et al., 2015).

The Perso-Arabic script support in Unicode is ever-growing, which is reflected by the number of recent proposals for new characters to better reflect the existing low-resource orthographies (Evans and Warren-Rothlin, 2018; Patel et al., 2019; Chitrali, 2020a,b) as well as to ease the encoding of the Quranic literature (Sh., 2022). The process of updating the standard is usually time-consuming, as demonstrated by the case of Torwali, which took two years from the time of the original proposal (Bashir et al., 2006) to encode the missing letter *hah with small arabic letter tah above* ⟨ݲ⟩ (U+0772). This letter completed the full character inventory for this emerging orthography in Unicode, which facilitated further developments of linguistic resources (Uddin and Uddin, 2019). As we found in our experiments, in the absence of the required characters, visually confusable variants or sequences of variants from foreign orthographies are often used by the Unicode-compliant input methods and converters from non-Unicode compliant fonts. Conversely, these methods take time to catch up with the Unicode standard once it introduces the missing features. To this one can add multiple confounding factors involved in the modern evolution of orthographies for hitherto unrecorded languages, which leads to rich orthographic diversity even among neighboring languages. For example, according to Bashir (2015, p. 14), the retroflex voiceless sibilant /ʂ/ present in several languages of northwestern Pakistan is represented differently by the regional writing systems: they all share the same *rasm* shape for letter *seen* ⟨س⟩ modified as: (1) *seen with small arabic letter tah and two dots* ⟨ݰ⟩ (U+0770) for Khowar; (2) *seen with extended arabic-indic digit four above* ⟨ݽ⟩ (U+077D) for Burushaski; (3) *seen with four dots above* ⟨ݜ⟩ (U+075C) for Torwali; (4) *seen with two dots vertically above* ⟨ݭ⟩ (U+076D) for Gowri; and (5) the corresponding two distinct characters for Kalasha and Shina remain unrepresented in Unicode.

| Nastaliq | Naskh | Transcription |
|---|---|---|
| تیرا غم ہے تو غم دہر کا جھگڑا کیا ہے | تیرا غم ہے تو غم دہر کا جھگڑا کیا ہے | terā ġham hai to ġham-e-dahr kā jhagḌā kyā hai |
| تیری صورت سے ہے عالم میں بہاروں کو ثبات | تیری صورت سے ہے عالم میں بہاروں کو ثبات | terī sūrat se hai aalam meñ bahāroñ ko sabāt |

Table 2: Urdu in Nastaliq (left), Naskh (center) and the corresponding transliteration (right). Samples taken from a poem (*nazm*) by Faiz Ahmed Faiz.

# 3 Perso-Arabic Script Diaspora: Selected Language Summaries

In what follows, we briefly delve into some adaptations of Perso-Arabic script. We limit our discussion to Perso-Arabic orthographies of eight languages, some of which are written in several scripts. The five languages from Indo-European family are Central Kurdish (Sorani), Kashmiri, Punjabi, Sindhi and Urdu. Two further languages, South Azerbaijani and Uyghur, come from the Turkic family. Finally, we provide a brief overview of the Perso-Arabic orthography called Jawi for Malay from the Austronesian family. The concise language-specific letter inventories are provided in Appendix A. Our software covers more orthographies, such as Balochi, Dari, Modern Standard Arabic, Pashto, Persian and Uzbek, yet we felt that our choice of the above eight writing systems is representative of the kinds of normalization challenges one is likely to encounter.

## 3.1 Urdu

Ethnologue classes Urdu as the tenth most spoken language in the world with over 70 million speakers using Urdu as their first language.[7] The national language of Pakistan, one of the 22 official languages of India, and a registered dialect in Nepal, Urdu is also spoken and used in 30 odd countries.[8]

The origins of Urdu are debatable and some scholars trace it back to the 6th century CE (Schmidt, 2007) but it was the Muslim invasion of Sindh in 711 which acted as a catalyst. By the time of the Mughal Empire and at the end of the 18th century it was the lingua franca around Delhi and was called *Zaban-e-Urdu*,[9] the word Urdu derived from the Turkic word *ordu* for "army" (Lelyveld, 1994).[10] The expansion of the Sultanate to the south gave rise to Dakhani Urdu (Mohamed, 1968). Urdu has a close association with Hindi since they share a common Indo-Aryan origin. Whereas Urdu is written in Perso-Arabic, Hindi uses Devanagari. The difference is best seen in the two versions of "The Chess Players" story written by Munshi Premchand, who authored the same story in both languages (Davis, 2015).[11]

Urdu as used in India and Pakistan is written in the Nastaliq style — a writing style developed in Iran from the Naskh style around the 13th century. Easy to write by hand, it posed problems when ported to metal type. Digital typography has to a certain extent solved the problem and text can be seen in the

[7] https://www.ethnologue.com/guides/ethnologue200
[8] https://www.ethnologue.com/language/urd
[9] Urdu was also called *Hindi*, *Hindavi*, *Hindustani*, *Dehlavi*, and *Lashkari*. But the term Urdu became most acceptable.
[10] https://www.rekhtadictionary.com/meaning-of-urdu
[11] https://thewire.in/culture/why-the-perso-arabic-script-remains-crucial-for-urdu

| Nastaliq | Naskh | Transcription |
|---|---|---|
| کافی آوو سکھی سہلیو مل مسلت گوئے | کافی آوو سکھی سہلیو مل مسلت گوئے | kaafī - aavo sakhī saheliyo mil maslat goīye |
| آپو آپنی گل نوں بھر ہنجھو روئے | آپو آپنی گل نوں بھر ہنجھو روئے | aapo aapnī gal nuuñ bhar hanjhū roīye |
| کھڈے لالچ لگیاں میں عمر گوائی | کھڈے لالچ لگیاں میں عمر گوائی | khaDe lālach lagyāñ maiñ umar gavā.ī |
| کدے نہ پونی ہتھ لے اک تندڑی پائی | کدے نہ پونی ہتھ لے اک تندڑی پائی | kade na puunī hath lae ik tandaḌī paa.ī |

Table 3: Shahmukhi in Nastaliq (left), Naskh (center) and the corresponding transliteration (right). Samples taken from a poem by Baba Farid.

Nastaliq style, however media on the Web prefers Naskh (Parhami, 2020). A sample of two lines of a *nazm* by Faiz Ahmed Faiz in Table 2 demonstrates the differences between the two styles.[12]

The Urdu writing system is an *abjad*, borrowed from Persian which in turn is borrowed from Arabic. Persian added four characters (⟨پ⟩, ⟨ژ⟩, ⟨گ⟩, and ⟨چ⟩) to the 28 basic characters borrowed from Arabic, bringing the total to 32. Persian further modified the character set by replacing the Arabic characters ⟨ي⟩ and ⟨ك⟩ with ⟨ی⟩ and ⟨ک⟩, respectively. To these 32 Persian characters, Urdu added: (1) the three letters ⟨ٹ⟩, ⟨ڈ⟩, and ⟨ڑ⟩ to accommodate retroflexes; (2) ⟨ں⟩ to handle nasalization; (3) the *two-eyed he* ⟨ھ⟩ to accommodate the aspirated forms of 17 or 18 letters; (4) the *yeh baree* ⟨ے⟩ to represent /e/ at the end of the words; and (5) ⟨ہ⟩ *gol he*, also called ***choṭī he***. Since diacritics are a systematic component of Perso-Arabic, this was possible without upsetting the graphic equilibrium of the script (Coulmas, 1999, p.560). The added *high hamza* placed above *farsi yeh*, *yeh baree*, *heh goal* and *waw* is used to create additional values and the *teh marbuta* ⟨ۃ⟩ marks feminine gender for nouns and adjectives. A further 5 characters were added to represent the 10 vowel phonemes, and an additional 5 to 10 diacritics were used when precision was needed.[13]

## 3.2 Punjabi (Shahmukhi)

The Shahmukhi writing system is used to record Punjabi in the Perso-Arabic script. According to Ethnologue this language is mainly spoken in Pakistan but also in other countries, especially in Punjab in India, with a total number of Punjabi speakers around 66 million.[14] The language is also known as Jangli, Lahanda, Lahnda, Lahndi, Panjabi, Panjabi Proper and Punjapi. Historically Shahmukhi was used by Sufi poets of the then Punjab region. One of the earliest instances of this writing system is its use by the Sufi poet of Punjab, Baba Farid in the 12th century (Singh and Gaur, 2009). After the partition of India, Shahmukhi became the writing system of choice for writing Punjabi by the Muslim population in Punjab. Hindus and Sikhs in the Indian state of Punjab adopted the Brahmic Gurmukhi script to write Punjabi, giving rise to Eastern Punjabi (Grewal, 2004). The relationship between Shahmukhi and Gurmukhi closely parallels that between Urdu and Hindi. Shahmukhi is an *abjad* and is written from right to left. It was highly influenced by Persian, but the present day writing system was modified to suit the requirements of the Punjabi language and as in the case of Urdu, a considerable number of characters were added. Like Urdu, Shahmukhi favours Nastaliq, but Naskh is used by digital media on the web. A sample of four verses from a poem by Baba Farid in both styles is shown in Table 3.

Urdu and Shahmukhi share the same character set, except that Shahmukhi admits a few more letters. The number of characters in Shahmukhi, like Urdu, is a matter of debate and some scholars admit four more characters ⟨ٻ⟩, ⟨ڄ⟩, ⟨ݙ⟩, and ⟨ڳ⟩ in addition to the retroflex lateral *lla* ⟨لؕ⟩ and the retroflex nasal *nna* ⟨ݨ⟩ (Bashir and Conners, 2019, pp.62,77). Of these, our analysis shows that only ⟨لؕ⟩ and ⟨ݨ⟩ seem to be in use. The ⟨ں⟩ character is used to mark end-of-word nasals. Like Urdu, the *two-eyed he* ⟨ھ⟩ is used to accommodate the aspirated forms of 17 or 18 letters. To these can be added the *high hamza* (U+0674) placed above *farsi yeh*, *yeh baree*, *heh goal* and *waw* to create additional values. Finally, five characters are used to represent the 10 vowel phonemes, and an additional 5 to 10 diacritics are used when precision is needed in cases such as consonant clusters or gemination. Shahmukhi and Urdu are thus mutually intelligible as writing systems.

[12] https://www.rekhta.org/nazms/

[13] The number of characters is debated, cf. https://www.dawn.com/news/919270.

[14] https://www.ethnologue.com/language/pnb

| Naskh | Devanagari | Transliteration |
|---|---|---|
| سَڄَڻُ ۽ ساڻيهُه ڪَنهِن اَڻاسي وِسري، | सज्रणु ऐं साणेहु कंहिं अणासी विसरी, | sajanu ain sanehu kanhin anasi visri, |
| حيفُ تِنين کي هوءِ، وَطَنُ جِنِ وِساري. | हैफ़ तनीं खे होइ, वतनु जिनि विसारी. | haif tanin khe hoi, vatanu jini visari. |

Table 4: Sample of Sindhi in Naskh (left), Devanagari (center), and the corresponding Latin transliteration (right). Samples taken from a poem by Shah Abdul Latif Bhittai.

## 3.3 Sindhi

Sindhi is an Indo-Aryan language spoken by the inhabitants of Sindh in the western part of the Indian subcontinent. It is one of the official languages of Pakistan and one of the 22 scheduled languages in India. Thanks to the Sindhi Diaspora it is spoken in quite a few countries and as per Ethnologue, has over 33 million speakers around the world.[15] Sindhi is recorded both in Perso-Arabic as well as Devanagari scripts. The traveler Al-Biruni in his Tarikh-al-Hind states that Sindhi was previously written in three scripts: Ardhanagari, Mahajani and Khudabadi (Sachau, 1910). But, the standardization of the Sindhi Perso-Arabic writing system ("arabi Sindhi") dates back to the 19th century. Prior to that, Sindhi Muslims had made attempts to write the language using Arabic, but the formal character set of Sindhi, as it is known today, goes back to 1853 when it was standardized by the British colonial authorities (Dow, 1976) and a set of 52 letters to accommodate the complexities of the sound system of Sindhi was identified. Sindhi is an *abjad* but unlike Urdu or Shahmukhi, Sindhi is only written using the Naskh style. A sample stanza from "Shah Jo Risalo" by Shah Abdul Latif Bhittai (Lajwani and Mirjat, 2021) shown in Table 4.

The addition of digraphs and the *hamza* over *yeh* and *waw*, as well as the diacritics to indicate the short vowels placed above *alef* and *waw*, brings the size of modern Sindhi letter inventory to 64 (Lekhwani and Lekhwani, 2014). For short vowels in particular, the following four letters ⟨اَ⟩, ⟨اِ⟩, ⟨اُ⟩, ⟨وُ⟩, composed by placing diacritic marks *fatha*, *kasra* and *damma* over *alef*, and the *damma* over *waw*, were added. To accommodate the large number of characters in its repertoire, Sindhi modified the Arabic *rasm* by addition of more *i'jām* dots.

Certain features of the character set of Sindhi make for the uniqueness of the writing system. Unlike Urdu or Shahmukhi, the *high hamza* is already accommodated over ⟨ؤ⟩ and ⟨ئ⟩. Sindhi admits four implosives ⟨ڳ⟩ *g̱a*, ⟨ڄ⟩ *j̱a*, ⟨ڏ⟩ *ḏa*, ⟨ٻ⟩ *ḇa*, and two single letter words ⟨۽⟩ *ain* ("and"), and ⟨۾⟩ *men* ("in"). Like Urdu, Sindhi has four letters to indicate /z/: ⟨ض⟩, ⟨ظ⟩, ⟨ز⟩, ⟨ذ⟩; three letters for /s/: ⟨س⟩, ⟨ص⟩, ⟨ث⟩; and two letters for /h/: ⟨ح⟩ and ⟨ه⟩. However, unlike Urdu which uses the *two eyed he* or *he do chashmi* ⟨ھ⟩ to mark the aspirates, Sindhi has individual characters for all the aspirates with the exception of ⟨گھ⟩ *gha*, ⟨جھ⟩ *jha*, and ⟨ڙھ⟩ *ṛha*. Vowel diacritics are not normally used. However if needed Sindhi has three diacritics used to indicate the short vowels. Additional diacritics are used to mark consonant clusters (*sukun*, `U+0652`) and gemination (*shadda*, `U+0651`).

## 3.4 Kashmiri

Kashmiri is a language from the Dardic family spoken in the Union Territory of Jammu and Kashmir, Himachal Pradesh and its outlying regions (Koul and Wali, 2015). Ethnologue lists around 7 million Kashmiri speakers in India and other countries.[16] It is a statutory language of provincial identity in Jammu and Kashmir[17] and is one of the 22 scheduled languages in India.[18] According to Kachru (2016), Kashmiri is the only Dardic language with a literary tradition and for which the written records have survived. Kashmiri is one of the three scheduled languages of India that is written using a Perso-Arabic script, the other two being Urdu and Sindhi. As in the case of Sindhi, Kashmiri is also written in Devanagari script, suitably modified to accommodate the sounds of the language.[19]

[15] https://www.ethnologue.com/language/snd

[16] `https://www.ethnologue.com/language/kas`

[17] In 2020, the Parliament of India passed a bill to make Kashmiri an official language of Jammu and Kashmir along with Dogri, Hindi, Urdu and English.

[18] `https://rajbhasha.gov.in/en/languages-included-eighth-schedule-indian-constitution`

| Nastaliq | Transliteration | Translation |
|---|---|---|
| کَنَوٕ کِن بَتہٕ لَدُن | Kanav kin batı ladun | Stuff rice through the ears: to overfeed |
| کَنَس بَتہٕ لَدُن | Kanas batı ladun | Stuff the ear with rice: advice wasted on a fool |

Table 5: A sample of Kashmiri proverbs.

Historically Kashmiri was written in the Sharada script, an *abugida* from the Brahmic family (Khaw, 2015). Sharada fell into disuse because it could not represent the complex sound system of the language. Successive invasions of the region slowly led to the adoption of the Arabic script. By the 14th century, Muslim rule in Kashmir was established and Kashmiri in Perso-Arabic script was adopted (Yatoo, 2012). The writing system evolved with time and the Arabic *rasm* were suitably adapted to add new characters to the repertoire. Today the Perso-Arabic script is recognized as the official writing system for the language. It is written in both Naskh and Nastaliq; and although the latter is favoured as the desired style, digital media prefers Naskh owing to the non-availability of a Nastaliq font for the script. Kashmiri is renowned for its proverbs (Koul, 2006; Kachru, 2021) and a sample of two proverbs[20] in Nastaliq, with the corresponding transliterations and English translations, is shown in Table 5.

All the vowel sounds of Kashmiri are regularly indicated in its orthography. This makes the writing system somewhat closer to an alphabet similar to Sorani Kurdish and Uyghur. The consonant inventory of Kashmiri consists of 37 letters. Some of these letters are shared with Urdu, Khowar and Shahmukhi orthographies, like the letter *rreh* ⟨ڑ⟩ for representing voiced retroflex flap /ɽ/ or the *ddal* ⟨ڈ⟩ for the voiced retroflex plosive /ɖ/. Kashmiri, like Urdu and Shahmukhi, uses the *two-eyed he* ⟨ھ⟩ in the construction of aspirated consonants. However, unlike Urdu and Shahmukhi only six such digraphs are permitted: ⟨تھ⟩, ⟨ٹھ⟩, ⟨پھ⟩, ⟨چھ⟩, ⟨ژھ⟩, and ⟨کھ⟩. The character ⟨ؠ⟩, a *yeh with a ring below* needs special mentioning. Kashmiri uses ⟨ؠ⟩ to mark palatalization which is a common feature in the language.

Kashmiri has one of the largest inventories of vowel letters, which are arranged in eight pairs of short and long vowels.[21] Kashmiri uses the *kasra*, *damma* and *fatha* as short vowel diacritic markers. Kashmiri modifies the *waw* to add new vocalic values: ⟨ۄ⟩ *waw with ring* to represent the sound /ɔ/; ⟨ۆ⟩ *waw with inverted v on top* for /o/; and ⟨ٗو ⟩ *waw with inverted damma* for a long /uː/. Additionally, *yeh baree* with an *inverted small v* marks the short /e/. Two combining marks are unique to Kashmiri, these are the *wavy hamza above* and *wavy hamza below*. The first is always used in conjunction with *alef* and represents a long schwa /əː/, while the second is used along with *alef* to represent /ɨː/.

Similar to Urdu and Shahmukhi, Kashmiri nasalisation is marked by the *noon ghunna* ⟨ں⟩ which can only occur in final position. When in medial position it is replaced by the letter *noon* ⟨ن⟩. Kashmiri also uses two other combining marks to mark gemination using *shadda*, and *sukun/jazm* (also called a *vowel killer*) to mark consonant clusters. The rendering of standard *sukun* diacritic (`U+0652`) is unique to Kashmiri writing system and has the shape of an inverted ⟨v⟩.

## 3.5 Central Kurdish (Sorani)

Sorani is the Perso-Arabic writing system used to write the Kurdish language in Iraq, mainly in Iraqi Kurdistan (Haig, 2018). This Indo-Iranian language is also spoken in regions adjoining Iran and Turkey. Ethnologue identifies three geo-linguistic variants of the language depending on where it is used: Central (Zimane Sorani), Northern (Kurmancî) and Southern (Kurdî Xwarîn or Pehlewaní). The name Sorani derives from the Soran Emirate, located in the area known today as Iraqi Kurdistan. Ethnologue lists around 4.7 million Sorani speakers in Iraq and total number of Sorani speakers in all countries at 5.3 million.[22] Unlike other languages in this study, Kurdish is not recognized today as the official language in any of the regions where it is used,[23] despite popular movements by Iraqi Kurds to give the language an official status.

[19] कॉशुर (Koshur)

[20] `https://kashmiridictionary.org/kanas-bati-ladun/`

[21] According to `https://r12a.github.io/scripts/arabic/ks.html`. Slightly different inventory is provided in `https://kashmiridictionary.org/category/learn-kashmiri/vowels-learn-kashmiri/`.

| Naskh | Latin |
|---|---|
| بیست ملیۆن کەس داوای بلیت دەکەن بۆ بەشداری کردن<br>لە دوا کۆنسێرتی لێد زێپالین | bîst miliyon kes dawayi bilît deken bo beşdarî kardin<br>lah dawa konisêrtî lêd zêpalîn |

Table 6: Sorani sentence in Naskh and Latin translating as "Twenty million people are asking for tickets to participate in the last Led Zeppelin concert".

Sorani in Perso-Arabic traces its origins to the Sulaimani region. The first trace of this writing system is found in "Mahdîname" by Mullah Muhammad Ibn ul-Haj completed circa 1762 (Bozarslan et al., 2021). The rise of the Baban dynasty encouraged the growth of Sorani and it became a medium for prose and poetry (Khalid, 2015). This continued until the Baban dynasty was overthrown around 1856. However, under British rule in the 19th century, Sorani literature and journalism flourished and multiple attempts were made to standardize the writing system, which led to eventual codification of the Sorani alphabet in the 1920s (Campbell, 1994).

Unlike most other Perso-Arabic writing systems, Sorani is a true alphabet, the vowels being explicitly marked. Sorani is written in Naskh style. A sample text from the Pewan corpus (Esmaili et al., 2013) is provided in Table 6.[24] As is the case with all languages adopting Perso-Arabic, in order to represent the phonemic features of the language, Sorani has evolved a system of letters some of which are unique to this writing system. These include the three unique consonant letters that are constructed by adding dots ⟨ڤ⟩ for /v/ or appending a small *v* below or above: ⟨ڕ⟩ for /r/, and⟨ڵ⟩ for /ɫ/. Similar to other Perso-Arabic writing systems, the vowel set borrows from the consonant set in that *yeh* and *waw* double as vowels and consonants. *Alef* is used as a vowel. The long /u:/ is indicated by doubling the *waw*. *Waw* and *yeh* with a small v above indicate /o/ and /e/, respectively.

Although the *kasra* is not part of the modern Sorani orthography, it is rarely used in some dictionaries for disambiguating certain pronunciations,[25] where it is used to mark a short /ɪ/ that is otherwise unrepresented in modern Perso-Arabic orthography.[26]

## 3.6 Uyghur

Uyghur, also written as Uighur, is a Turkic language spoken in the region in and around what is known as the Xinjiang Uyghur Autonomous Region in Northwest China.[27] Due to a politically and culturally motivated diaspora, Uyghur is spoken in Turkic countries such as Kazakhstan, Kyrgyzstan, Uzbekistan and Turkey, but also by the smaller Uyghur migrant communities elsewhere (Dillon, 2009). Ethnologue estimates the number of Uyghur speakers at around 10 million in China and total speakers in all countries at around 10.4 million. The modern Uyghur writing system should not be confused with Old Uyghur which was written using Sogdian script (Wilkens, 2016). Historically the writing system dates back to the 10th century when the Perso-Arabic script was introduced along with the spread of Islam and which evolved after considerable changes over the centuries into what is recognized as the modern Uyghur Perso-Arabic orthography (Brose, 2017). The writing system for the language underwent extensive changes, including being changed to the Cyrillic and Latin scripts and even the Pinyin romanization system for political reasons. It was not until 1982 that the Arabic Uyghur alphabet was reinstated (Dwyer, 2005). As of today the language has four writing systems: Uyghur Arabic used in the Xinjiang province of China, Uyghur Cyrillic in Kazakhstan, Uyghur Latin in Turkey and Uyghur Pinyin, which is not used much (Hamut and Joniak-Lüthi, 2015).

[22] https://www.ethnologue.com/language/ckb
[23] In 2006, Duhok Governorate began using Kurmanji as their official language as a way of resisting Sorani.
[24] https://sinaahmadi.github.io/resources/pewan.html
[25] Private correspondence from Aso Mahmudi (2022).
[26] According to Ahmadi (2019, §2.2, p.3), the corresponding letter of Latin-based orthography of Kurmanji dialect is ⟨i⟩.
[27] https://www.ethnologue.com/language/uig

| Naskh | Latin |
|---|---|
| پرەزىدەنت ماراسىم قاتناشقۇچىلىرىنى ۋاتان قىمايىچىسى كۈنى<br>ۋا غالىبىيات كۈنى بىلان تابرىكلاپ قاربيي كھىزماتچىلارنىنھ<br>تامىنلاشكا ئالاقىدا قاسسا قوشۇۋاتقانلىغىنى ئاتاپ كورساتتىن | prezident marasim qatnashquchilirini vatan qimayichisi kuni<br>va ghalibiyat kuni bilan tabriklap qarbiy khizmatchilarninh<br>taminlashka alaqida qassa qoshuvatqanlighini atap korsattin |

Table 7: Uyghur sample in Naskh transliterated from Latin text.

| Nastaliq | Naskh |
|---|---|
| میرزا شفیع واضح آذربایجان شاعیری و موتفکّیری، معاریفچی و پداقوق. | میرزا شفیع واضح آذربایجان شاعیری و موتفکّیری، معاریفچی و پداقوق. |

Table 8: A line from South Azerbaijani Wikipedia: "Mirza Shafi Vazeh – Azerbaijani poet, thinker, enlightener and teacher".

Unlike most other writing systems using Perso-Arabic, but like Sorani, Uyghur writing system is an alphabet, i.e., the vowels are explicitly marked. Uyghur is written in Naskh style, a sample of which is shown in Table 7.[28] As is the case with all languages which have adopted the Perso-Arabic script, in order to represent the phonemic features of the language, Uyghur writing system has evolved an original repertoire of letters. Apart from the letters borrowed from the original Arabic script, four letters are derived from Persian writing system: ⟨پ⟩, ⟨چ⟩, ⟨ژ⟩, ⟨گ⟩; the ⟨ڭ⟩, which represents a velar nasal, common to Turkic languages, is derived from the Arabic *kaf* ⟨ك⟩ with three dots positioned above the letter. The *two-eyed he* ⟨ھ⟩ is also used, similar to Kazakh, Urdu, Sindhi and Shahmukhi among other languages.

Extensive use of the *waw* is made, which is modified in productive ways to represent the vowels: ⟨ۈ⟩ with a superscript *alef* to represent the sound /yu/, ⟨ۆ⟩ with a *small v* on top to represent a front rounded vowel /ø/, ⟨ۇ⟩ with a *damma* on top for a long /u:/ and ⟨ۋ⟩ with *three dots* on top represents the semivowel /w/. Uyghur uses the Arabic *yeh* ⟨ي⟩ for the semivowel /j/, the *alef maqsura* ⟨ى⟩ for the /i/ and ⟨ې⟩ *yeh* with two dots below to represent /e/. Additional combining marks are used to mark consonant clusters (Arabic *sukun*, `U+0652`) or gemination (Arabic *shadda*, `U+0651`).

## 3.7 Southern Azerbaijani

Azerbaijani, also known as Azeri, Azari, Azeri Turkish and Azerbaijani Turkish, belongs to the Turkic language family, more specifically to the Western Oghuz branch (Mokari and Werner, 2017). It is spoken by over 23 million people, mainly in Azerbaijan, Iran, Georgia, Russia and Turkey, and also in Iraq, Syria and Turkmenistan.[29] Two varieties of the language are recognized: Northern Azerbaijani and Southern Azerbaijani. Northern Azerbaijani is spoken in the Republic of Azerbaijan, where it is the official language. Southern Azerbaijani spoken by around 14.6 million people is confined to the northwest of Iran and is often called *Turki* (تورکی).[30] Due to migrations and trade it is also used in parts of Iraq and Turkey, and in Afghanistan and Syria. Whereas Northern Azerbaijani uses either the Cyrillic script (in Dagestan) or Latin (the official script in Azerbaijan), Southern Azerbaijani uses the Perso-Arabic script. The Naskh style is favoured in day to day use but Nastaliq is sometimes used, mainly for book titles and also for handwriting, as demonstrated in Table 8.

Historically Old Azeri (*Āḏarī*) was the Indo-Iranian language spoken in Persian Azerbaijan before the arrival of the Turkic-speaking populations to the region (Yarshater, 2011). The language was gradually replaced with Turkish as the migration of Turkic speakers increased and by the 18th century Turkish was recognized as the language of Azerbaijan, although the name *Āḏarī* was retained as Azeri. The traces of the original *Āḏarī* language can still be found in modern Azeri today (Bosworth, 2011). The arrival of the muslim Turkish speakers in South Azerbaijan was accompanied by the Perso-Arabic *abjad* which became the official script of Azerbaijan until the 1920's, when, for political reasons, competing Cyrillic and Latin scripts entered the scene (Hatcher, 2008). The Azerbaijani Perso-Arabic writing system saw considerable mutation over the centuries: 28 letters (all from Arabic) initially, increased to 32 letters with additions from Persian and, finally, 33 letters due to an addition from the Ottoman Turkish. None of these solutions were found suitable for Azerbaijani and reforms were proposed during the 19th and 20th centuries which finally created the character set of Southern Azerbaijani as it is known today.[31]

[28] The Latin text obtained from UygurAvazi newspaper (`https://uyguravazi.kazgazeta.kz/`) was converted to Uyghur through a script converter from `http://www.elipbe.com`.

[29] `https://www.ethnologue.com/language/aze`

[30] `https://www.ethnologue.com/language/azb`

| Naskh | Rumi |
|---|---|
| ينله ݢراڠن سواتو ماده مڠارڠكن شعير ترلالو ايندە، | inilah gerangan suatu madah mengarangkan syair terlalu indah, |
| ممبتولي جالن تمڤت برڤينده، | membetuli jalan tempat berpindah, |
| د سانله اعتيكات دڤربتولي سوده | di sanalah i'tikat diperbetuli sudah |

Table 9: Sample of Jawi in Naskh (left) and Rumi (right).

The modern inventory consists of 42 letters. The majority of letters are borrowed from the Arabic and Persian orthographies. Nine letters (⟨ذ⟩, ⟨ژ⟩, ⟨ص⟩, ⟨ض⟩, ⟨ط⟩, ⟨ظ⟩, ⟨ع⟩, ⟨ح⟩, and ⟨ث⟩) are exclusively used for spelling Persian and Arabic loanwords and names. An extra letter *keheh with three dots above* ⟨ݣ⟩ is used to indicate the voiced velar nasal /ŋ/, similar to Uyghur (Daniels, 2014, p. 31). Like all Turkic languages, Azerbaijani has a rich vowel system (Johanson and Csató, 2021). Three core shapes ⟨ا⟩, ⟨و⟩, and ⟨ی⟩, modified with various diacritics, form the letters of the vowel set. Letter ⟨ئ⟩ represents the sound /e/, ⟨ىٚ⟩ the unrounded back vowel /ɯ/ and ⟨ي⟩ represents /iː/. The *rasm* for *waw* is adapted in four ways. Apart from the intrinsic value of *waw*, ⟨ۇ⟩ is used for /uː/, ⟨ۆ⟩ for the front open rounded vowel /y/, ⟨ؤ⟩ for the front rounded vowel /ø/ and the digraph *waw with sukun* ⟨وْ⟩ represents /o/. The letter ⟨ه⟩ is used only in the final form to mark the diphthong /ae/. In addition, Azerbaijani orthography admits three combining marks, *fatha*, *damma* and *kasra*, to mark short vowels, and also additional diacritics to mark consonant clusters (*jazm/sukun*, U+0652) or gemination (*shadda*, U+0651).

## 3.8 Malay (Jawi)

Jawi is a Perso-Arabic writing system used for recording the Malay language from the Austronesian family and several other languages of Southeast Asia (Kratz, 2002). With the advent of Islam in Southeast Asia around the 14th century, the Pallava script, Nagari, and old Sumatran scripts which were used in writing Malay, were replaced by the Perso-Arabic script and by the 15th century Jawi had spread to Brunei, Indonesia and even Thailand due to trading (Coluzzi, 2020). Its dominance remained till the 20th century when Jawi was replaced by the Latin script (*Rumi*) and was confined to religious and cultural rituals. Today apart from Malaysia, Jawi has the status of an official writing system in Brunei and also in Indonesia, where Jawi has been assigned a regional status (Abdullah et al., 2020). Unlike Urdu or Shahmukhi, and, to a lesser extent Persian, Jawi favours the Naskh style, demonstrated by the sample quatrain from "Syair Perahu", a Sumatran Sufi poem (Braginsky, 1975), in Table 9.

In addition to the 28 basic characters from Arabic,[32] Jawi added extra characters to suit its requirements and introduced the following: ⟨چ⟩ *ca*, ⟨ڠ⟩ *nga*, ⟨ڤ⟩ *pa/fa*,[33] ⟨ݢ⟩ *ga*, and ⟨ڽ⟩ *nya*. The letter ⟨ۏ⟩ *v* was added for representing foreign loanwords.[34] This brings the size of modern Jawi inventory to 37 letters (DBP, 2006). In addition, three more characters are possible due to the adjunction of the *high hamza* ⟨ء⟩ above *alef* ⟨أ⟩, below *alef* ⟨إ⟩, and above *yeh* ⟨ئ⟩ (MS, 2012).

As in Arabic, vowel diacritics are not normally used. However, if needed Jawi has three diacritics used to indicate the short vowels: *fatha*, *damma*, and *kasra*. A major feature of the language is the use of full reduplication of the base word (Prentice, 1990). This is represented in Jawi with the Arabic numeral ⟨٢⟩ ("2") as in انجيڠ٢ *anjeng-anjeng* ("dogs") as a shorthand for the equivalent longer spelling انجيڠ-انجيڠ for the plural form of a noun انجيڠ /andʒeŋ/ ("dog").

[31]For example, in modern Azerbaijani, letter *keheh* ⟨ک⟩ has replaced the older Arabic *kaf* ⟨ك⟩.

[32]Some scholars, like R. O. Windstedt, believe that Jawi borrowed the characters from the Persian, rather than Arabic, orthography (Winstedt, 1961).

[33]The letter *fa* ⟨ڤ⟩ was used to represent *pa* because the sound /f/ does not exist in Malay and was pronounced as /p/.

[34]The letter *va* ⟨ۏ⟩ is mostly used to spell English loanwords, e.g. اونيۏرسيتي ("universiti").

| Operation Type | FST | Language-dependent | Includes |
|---|---|---|---|
| NFC | $\mathcal{N}$ | no | – |
| Common Visual | $\mathcal{V}_c$ | no | $\mathcal{N}$ |
| Visual | $\mathcal{V}$ | yes | $\mathcal{V}_c$ |
| Reading | $\mathcal{R}$ | yes | $\mathcal{V}$ |

Table 10: Summary of script transformation operations.

## 4 Finite-state Transformations of Perso-Arabic Script

Below we provide a brief overview of the design for Perso-Arabic script normalization framework provided by the open-source Nisaba software package.[35] The design was partially inspired by prior formal approaches to computational modeling of Brahmic alphasyllabaries (Datta, 1984; Sproat, 2003) and, in particular, our prior work at Brahmic script normalization (Johny et al., 2021; Gutkin et al., 2022b). These approaches exploit the inherent structure which manifests itself in all the Brahmic *abugidas* in the notion of "orthographic syllable" or *akṣara* (Bright, 1999; Fedorova, 2012). In contrast to various Brahmic scripts, the Perso-Arabic *abjad* does not offer the same rigid orthographic structure. Nevertheless, a similar in nature formal approach to script normalization, designed to address the kind of Perso-Arabic script representation ambiguities outlined in §2, can be pursued. We previously showed that scripts, such as Thaana, that borrow their features from both script families are amenable to such formal analysis (Gutkin et al., 2022b).

Our script processing pipeline consists of multiple components implemented as finite-state grammars using Pynini (Gorman, 2016; Gorman and Sproat, 2021), which is a Python framework for compiling grammars expressed as strings, regular expressions, and context-dependent rewrite rules into (weighted) finite-state transducers (FSTs). The resulting FSTs can then be efficiently combined together in a single pipeline in a variety of downstream applications (Mohri, 1996, 2009). These component FSTs are shown in Table 10 and described below.

**Unicode Normalization** There exist language-agnostic procedures—part of the Unicode standard—that normalize text with Perso-Arabic string encodings to visually equivalent canonical normal forms. Normalization Form C (NFC) is a well-known and widely-used standard of this sort, and its application results in an equivalence class of visually identical strings that are all mapped to a single conventionalized representative of the class (Whistler, 2021). In the Nisaba library, the NFC standard is operationalized by compiling the transformations into an FST, which we denote as $\mathcal{N}$ in Table 10. The transformations include compositions and re-orderings, along with combinations of multiple such transformations.

Composition transformations can be illustrated with the following concrete example. The *alef with madda above* letter ⟨آ⟩ has two visually identical possible encodings: with two characters by adjoining *maddah above* to *alef* ({ `U+0627`, `U+0653` }), or as the single character that already includes the maddah (`U+0622`). The FST $\mathcal{N}$ transforms the two character encoding into the single character encoding, which does not change the appearance of the letter. Re-ordering transformations address multiple encodings that can arise with Arabic combining marks. As a concrete example, *shadda* (`U+0651`) followed by *kasra* (`U+0650`) yields the same rendering as *kasra* (`U+0650`) followed by *shadda* (`U+0651`). The NFC canonical form is the latter, hence the $\mathcal{N}$ FST transforms the former encoding to the latter. The string { *alef* (`U+0627`), *superscript alef* (`U+0670`), *maddah above* (`U+0653`) } is an example that transforms via $\mathcal{N}$ with both composition and re-ordering to the visually identical form { *alef with madda above* (`U+0622`), *superscript alef* (`U+0670`) }.

As noted above, $\mathcal{N}$ is language-agnostic, meaning that its transformations (taken from the NFC standard) do not violate any language's writing system rules.

**Visual Normalization** We use the term *visual* normalization—initially introduced in the context of Brahmic script normalization (Johny et al., 2021)—to denote transformations that are not part of NFC but that also result in canonical forms that are visually identical to the input. This is implemented via two FSTs, one for language-agnostic transformations and one language-specific, which are combined (via FST composition) with NFC into a single language-dependent FST: $\mathcal{V} = \mathcal{N} \circ \mathcal{V}_c \circ \mathcal{V}_l$, where $\circ$ denotes the composition operation (Mohri, 2009).[36]

[35] For more detailed treatment of this software please see Gutkin et al. (2022a).

| Kind of rewrite | FST | Letter | Variant (source) | Canonical (target) |
|---|---|---|---|---|
| position-independent | $\mathcal{V}_l^*$ | ⟨ڑ⟩ | *reh + small high tah* | *rreh* |
| non-final | $\mathcal{V}_l^{\mathrm{n}}$ | ⟨ک⟩ | *kaf* | *keheh* |
| word-final | $\mathcal{V}_l^{\mathrm{f}}$ | ⟨ى⟩ | *alef maksura* | *farsi yeh* |
| isolated-letter | $\mathcal{V}_l^{\mathrm{i}}$ | ⟨ه⟩ | *heh* | *heh goal* |

Table 11: Example Urdu components included in language-specific FST $\mathcal{V}_l$.

The language-agnostic FST, $\mathcal{V}_c$, consists of the small set of normalizations not included in NFC that apply to all supported languages. As a concrete example of this class of transformations, the two-character encodings of *waw* (U+0648) followed by either *damma* (U+064F) or *small damma* (U+0619) are mapped to *u* (U+06C7). Perso-Arabic "presentation forms" from Unicode Block A, which include ligatures and contextual forms for letter variants required by the writing systems for Persian, Urdu, Sindhi and Central Asian languages,[37] are also normalized to visually identical canonical forms by $\mathcal{V}_c$, as specified by Unicode NFKC normalization (Whistler, 2021). For example, letter *beeh isolated form* ⟨ﭒ⟩ (U+FB52) is normalized to *beeh* (U+067B), which is visually identical. The character *ligature lam with alef isolated form* ⟨ﻻ⟩ (U+FEFB) is transformed to two characters: *lam* ⟨ل⟩ (U+0644) followed by *alef* ⟨ا⟩ (U+0627).

Language-specific visually-invariant transformations, included in the FST denoted as $\mathcal{V}_l$, include four special cases related to positions in the word: word-final, non-final (i.e., word-initial and word-medial), isolated-letter and position-independent transformations. Each of these are compiled into their own FST, as shown in Table 11, then composed into a single $\mathcal{V}_l = \mathcal{V}_l^{\mathrm{i}} \circ \mathcal{V}_l^{\mathrm{f}} \circ \mathcal{V}_l^{\mathrm{n}} \circ \mathcal{V}_l^*$. Table 11 additionally presents some example transformations of each type, taken from the set of transformations required for Urdu.

**Reading Normalization** Gutkin et al. (2022b) noted the need for some additional normalization beyond those preserving visual identity for the Brahmic scripts, which they termed *reading* normalization. We also include this class of normalizations for Perso-Arabic, which we compile into the FST denoted $\mathcal{R}$ in Table 10. Full reading normalization is the finite-state composition of visual normalization with language-specific reading normalization: $\mathcal{R} = \mathcal{V} \circ \mathcal{R}_l$. For example, Persian, Shahmukhi, Kashmiri, Urdu and Sorani Kurdish all map from *yeh* ⟨ي⟩ (U+064A) to *farsi yeh* ⟨ی⟩ (U+06CC), while Uyghur, Sindhi and Malay employ the inverse of this transformation, as dictated by their respective orthographies.

# 5 Experiments

While outlining the potential issues that may arise with text written in the Perso-Arabic script is important, it is also useful to assess how common the issues may be in real-world text. To that end, we devised some experiments that derive natural language models from collected text and validate their quality both with and without normalization. If the phenomena being normalized are rare, then the difference between the conditions will be small; and if the normalizations do not result in better text representations, then the normalized conditions may exhibit a lower quality in the validation. In this section, we present the details of our assessment, first for statistical language modeling, which provides an intrinsic validation of model quality, followed by machine translation, which provides an extrinsic validation. As was mentioned in the introduction, the code for the experiments and the corresponding results for both validation types have been released (see Footnote 4 on page 2).

[36] Johny et al. (2021) provides details regarding composition and other operations used by FSTs in these normalizers.
[37] https://www.unicode.org/charts/PDF/UFB50.pdf

## 5.1 Language Modeling Experiments

For language modeling experiments we use Wikipedia data for eight languages: five Indo-European — Kashmiri, Kurdish (Sorani), Punjabi, Sindhi, and Urdu; Malay from an Austronesian group; and Uyghur and Azerbaijani from the Turkic group. A brief overview of experimental methodology is given in §5.1.1. The dataset preprocessing details are provided in §5.1.2. The details and results of statistical language modeling experiments can be found in §5.1.3.

### 5.1.1 Methodology

Language models are trained to predict the next token in a sequence given the previous tokens. Tokens can be variously defined as characters or words, or even as morepheme-like sub-word multi-character tokens. The intrinsic quality of the language model can be measured via the probability the language model assigns to attested exemplars, i.e., real text. The higher the aggregated probability of the attested text, the better the language model. See, e.g., Rosenfeld (2000) for more details on this long-standing validation paradigm. For this work, a key consideration is *comparability* – we need to ensure that models have access to the same training data and that the validation data is identical.

To ascertain whether script normalization has any significant impact on language model quality we follow a simple methodology. We adopt a $k$-fold cross-validation design where, for each language, we randomly shuffle the dataset and split it into the 80% training and 20% test folds, repeating the process $k$ times, where $k = 100$. At each iteration we train the models (e.g., language models as in §5.1.3) and evaluate them by computing the corresponding metric (e.g., cross-entropy).

Following the above procedure, statistics are assembled using $k$ observations for the baseline configurations that correspond to the original text and the actual testing configurations corresponding to the normalized text. Crucially, we start by generating the normalized data, recording all the sentences which contain the actual diffs in set $D$. During the generation of the training and test data for the baseline and testing configurations, we make sure that the sentences in $D$ are confined to the training set. In other words, we make sure that all the actual rewrites are confined to the training data for all the $k$ folds, which ensures that the test folds are always identical for normalized and unnormalized conditions. Given the baseline and the test metric distributions, we employ significance testing to validate the null hypothesis that the two distributions are identical; in other words, that the normalization has no significant impact on the model performance.

Three types of statistical hypothesis tests are used here. Assuming that the two groups are normally distributed an obvious choice is the two-sample (independent) $t$-test for comparing the means of the two populations (Zabell, 2008). Making an additional assumption that the population variances are not equal, we employ Welch's formulation of $t$-test (Welch, 1947) with Satterthwaite's degrees of freedom (Satterthwaite, 1946), referred to below as Welch-Satterthwaite (WS) test. The test provides the $t$ statistic, the $p$-value and the estimated confidence interval (CI) $[L, H]$ for the 95% confidence level at the significance level of $\alpha = 0.05$.

In addition, two non-parametric approaches are used here. A Mann-Whitney (MW) test (Mann and Whitney, 1947) and a more recent Brunner-Munzel (BM) test (Brunner and Munzel, 2000). Both tests provide the $t$ statistic and the $p$-value. The rationale for using multiple hypothesis tests is to see whether they all agree with other providing additional weight to the null (or alternative) hypothesis.[38]

### 5.1.2 Corpus Preprocessing

The process of preparing the Wikipedia data is kept simple. The datasets for each language are downloaded in the MediaWiki XML format. The particular version of the dump is restricted to the pages with their current versions including the metadata.[39] The key difficulty lies in extracting the actual plain text in native language from the structured XML data while weeding out the metadata. We use the `mwxml`

[38] All the algorithms are provided by the open-source `scipy.stats` and `statsmodels` packages.

[39] For example, a reasonably recent dump for Punjabi (Shahmukhi) is available here.

| Language | Code | $\beta$ | $N_l$ | $N_w$ | $N_l^r$ | $R_l(\%)$ | $N_w^r$ | $R_w(\%)$ |
|---|---|---|---|---|---|---|---|---|
| Kashmiri | ks | 0.6 | 3266 | 9721 | 530 | 16.22 | 442 | 4.55 |
| Kurdish (Sorani) | ckb | 0.8 | 839 750 | 794 475 | 42 437 | 5.05 | 67 569 | 8.5 |
| Malay | ms | 0.0 | 102 311 | 200 052 | 72 645 | 71.0 | 25 854 | 12.92 |
| Punjabi (Shahmukhi) | pnb | 0.8 | 1 075 820 | 886 399 | 145 391 | 13.51 | 16 443 | 1.86 |
| Sindhi | sd | 0.8 | 201 345 | 240 591 | 138 839 | 68.96 | 64 006 | 26.6 |
| South Azerbaijani | azb | 0.8 | 1 638 622 | 735 986 | 60 094 | 3.67 | 23 834 | 3.32 |
| Uighur | ug | 0.1 | 110 344 | 376 307 | 3600 | 3.26 | 19 461 | 5.17 |
| Urdu | ur | 0.9 | 3 595 095 | 799 610 | 6600 | 0.18 | 3632 | 0.45 |

Table 12: Details of preprocessed Wikipedia datasets.

Python package developed by the Wikipedia foundation to iterate over the articles in MediaWiki XML dump.[40]

For each article, we use the MediaWiki Parser from Hell package to parse the current revision of article's text.[41] Once the parse is complete, we strip the contents of all the "unprintable" content, such as templates, using the API provided by the `mwparserfromhell` package, and split the text by newlines. A simple script detection and filtering algorithm is used to decide whether to keep the sentence or drop it from the resulting data.[42] Any given $l$-character long sentence is dropped from the data if it contains less than $\beta \cdot l$ characters in native (Perso-Arabic) script, where $\beta \in [0, 1)$. The filtering factor $\beta$ is language-specific and is determined by informally examining the data.[43] This filtering process is crude in that it excludes any control for sentence or token length.

The preprocessing details are shown in Table 12. Each language is shown along with its Wikipedia code, the script filtering factor $\beta$ described above, the number of resulting lines $N_l$ and the number of unique tokens $N_w$. For the corresponding normalized text, $N_l^r$ denotes the number of lines that contain diffs and $N_w^r$ is the number of unique tokens that differ from the unnormalized version. The ratios between modified lines and token types are denoted $R_l$ and $R_w$, respectively. The tokenization process is relatively crude and involves splitting on the whitespace completely disregarding other types of punctuation, such as Perso-Arabic punctuation symbols. We refer to the output of tokenization as tokens, rather than words, because the data is quite noisy even after filtering.

When normalizing text we apply the Nisaba *reading* normalization grammar (Gutkin et al., 2022b), which subsumes all the grammars providing visual invariant transformations, i.e., NFC and *visual* normalization (Johny et al., 2021), as well as transformations that change the visual appearance of the Perso-Arabic tokens. According to Table 12, the normalization effects vary across languages. For Urdu, which is the largest dataset, the percentage of modified lines and token types is below one percent. This reflects the relatively low number of transformations currently enabled in the Nisaba Urdu grammars compared to the other six languages. The highest proportion of modified lines and tokens happens in Sindhi and Malay, while for Kashmiri (the smallest datasets) and Punjabi (Shahmukhi, the second largest) the number of modifications is relatively low. A description of how these modifications affect model quality follows next.

### 5.1.3 Statistical Language Models

For building $n$-gram language models we use the KenLM toolkit (Heafield, 2011)[44] which is fast and easy to use relative to alternatives. We used modified Kneser-Ney modeling options, as recommended (Heafield et al., 2013). In what follows, the terminology introduced in §5.1.1 is used. The experiments with character and word $n$-gram models are described in §5.1.3 and §5.1.3, respectively.

For a single fold, the criterion for splitting into training and test sets is to use the number of lines in the corpus. As a result, the number of training and test tokens (whether these are individual characters

[40] https://www.mediawiki.org/wiki/Mediawiki-utilities/mwxml
[41] https://github.com/earwig/mwparserfromhell
[42] We previously implemented a similar script detection algorithm for the Wikipron project.
[43] For Malay we use $\beta = 0$, i.e., no filtering.
[44] https://github.com/kpu/kenlm

| Language | Train | | Test | |
|---|---|---|---|---|
| | $\mu$ | $\sigma$ | $\mu$ | $\sigma$ |
| azb | 148 133 643.5 | 104 532.9 | 32 841 396.5 | 104 532.9 |
| ckb | 141 088 021.3 | 118 118.6 | 32 419 200.7 | 118 118.6 |
| ks | 252 357.7 | 3438.0 | 63 370.3 | 3438.0 |
| ms | 16 981 259.9 | 9569.6 | 3 394 367.1 | 9569.6 |
| pnb | 216 833 075.2 | 248 147.7 | 54 209 912.8 | 248 147.7 |
| sd | 36 855 104.9 | 95 212.8 | 9 212 888.1 | 95 212.8 |
| ug | 33 813 243.4 | 109 518.9 | 8 096 172.6 | 109 518.9 |
| ur | 368 794 342.1 | 240 334.8 | 92 205 182.9 | 240 334.8 |

Table 13: Character-level statistics for $k$-fold configurations and $m$ $n$-gram orders ($k = 100$, $m = 8$).

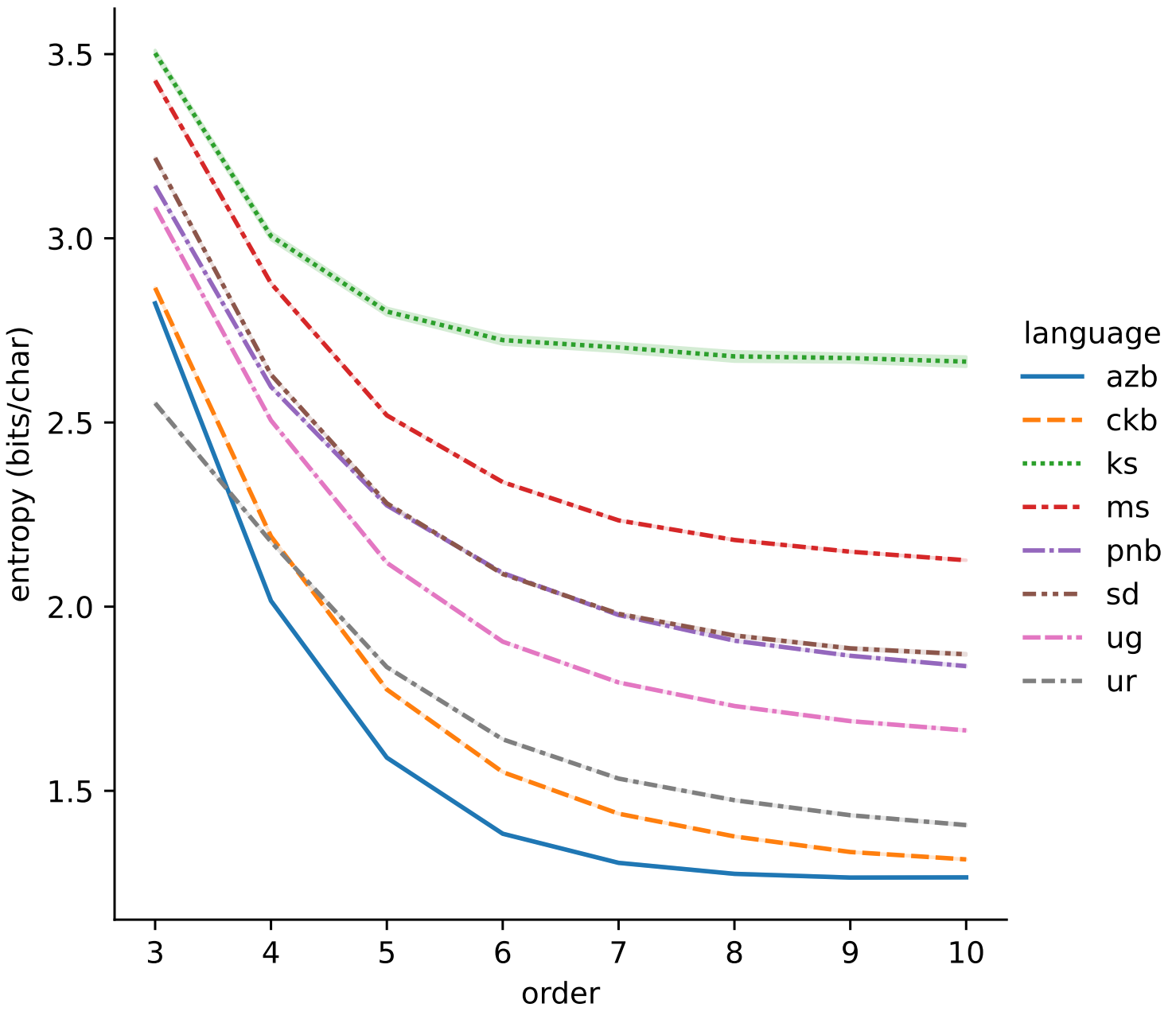

Figure 2: Average values of entropy (bits per character) over 100 runs vs. character $n$-gram orders.

or words) differ across the folds.

**Character Models** For each language and each of the $k = 100$ train/test folds we build $n$-gram character language models for orders $n \in [3, \ldots, 10]$. The character-level statistics computed for each language over all the folds and all the orders (amounting to 800 observations per language) are shown in Table 13, where the means and standard deviations are shown for the training and test datasets. Since the corpora are split by the number of lines, the resulting variances for character datasets are quite high.

The resulting cross-entropies (in bits per character) for the models built in this way from the unnormalized text are shown in Figure 2, where each point for each $n$-gram order in the curve is shown along with its corresponding error band computed over $k$ models. The plot for Kashmiri, the smallest dataset among the four languages, stands out in that the error band is clearly visible, especially for the higher orders for which the model overfits the training data. The plots for the rest of the languages show very low variance at each point in the plot for all the $n$-gram orders.

For each of the languages and each of the $n$-gram orders, we perform statistical hypothesis testing for the differences in mean cross-entropies between the character language models trained on the original (baseline, denoted $B$) and the normalized (test, denoted $T$) text for all the $k$ folds ($k = 100$). As mentioned in §5.1.1, for each fold, the sentences that contain (for $T$) or act as source of (for $B$) normalization diffs are kept in the training portion of the data. Full results for each of the languages are presented in Appendix B, and key summarizing values are shown for all in Table 14.

| Language | Measure | *n*-gram order 3 | 4 | 5 | 6 | 7 | 8 | 9 | 10 |
|---|---|---|---|---|---|---|---|---|---|
| South Azerbaijani | $\Delta_\mu$ | −0.012 | −0.004 | −0.002 | −0.001 | −0.002 | −0.002 | −0.002 | −0.002 |
| | % | 0.418 | 0.186 | 0.131 | 0.093 | 0.14 | 0.143 | 0.13 | 0.171 |
| | $p$ | 0.0 | 0.0 | 0.002 | 0.048 | 0.012 | 0.013 | 0.02 | 0.004 |
| Kurdish (Sorani) | $\Delta_\mu$ | −0.01 | −0.003 | −0.004 | −0.003 | −0.004 | −0.006 | −0.005 | −0.005 |
| | % | 0.36 | 0.16 | 0.24 | 0.2 | 0.28 | 0.44 | 0.43 | 0.39 |
| | $p$ | 0.0 | 0.0 | 0.0 | 0.0 | 0.0 | 0.0 | 0.0 | 0.0 |
| Kashmiri | $\Delta_\mu$ | −0.003 | −0.021 | −0.006 | −0.006 | −0.005 | −0.016 | −0.022 | −0.028 |
| | % | 0.08 | 0.71 | 0.21 | 0.22 | 0.2 | 0.63 | 0.85 | 1.07 |
| | $p$ | 0.647 | 0.007 | 0.544 | 0.515 | 0.564 | 0.075 | 0.014 | 0.003 |
| Malay (Jawi) | $\Delta_\mu$ | −0.065 | −0.062 | −0.06 | −0.064 | −0.067 | −0.07 | −0.07 | −0.07 |
| | % | 1.818 | 2.036 | 2.232 | 2.526 | 2.736 | 2.885 | 2.931 | 2.922 |
| | $p$ | 0.0 | 0.0 | 0.0 | 0.0 | 0.0 | 0.0 | 0.0 | 0.0 |
| Punjabi (Shahmukhi) | $\Delta_\mu$ | −0.011 | −0.013 | −0.01 | −0.008 | −0.007 | −0.006 | −0.007 | −0.007 |
| | % | 0.32 | 0.44 | 0.39 | 0.34 | 0.33 | 0.3 | 0.36 | 0.34 |
| | $p$ | 0.0 | 0.0 | 0.0 | 0.0 | 0.0 | 0.0 | 0.0 | 0.0 |
| Sindhi | $\Delta_\mu$ | 0.001 | 0.011 | −0.018 | −0.028 | −0.016 | −0.024 | −0.015 | −0.014 |
| | % | −0.03 | −0.28 | 0.47 | 0.79 | 0.46 | 0.71 | 0.45 | 0.42 |
| | $p$ | 0.479 | 0.0 | 0.001 | 0.0 | 0.001 | 0.0 | 0.007 | 0.015 |
| Uyghur | $\Delta_\mu$ | −0.002 | −0.001 | −0.004 | −0.004 | −0.003 | −0.004 | −0.004 | −0.005 |
| | % | 0.074 | 0.051 | 0.203 | 0.219 | 0.164 | 0.225 | 0.24 | 0.295 |
| | $p$ | 0.0 | 0.025 | 0.0 | 0.0 | 0.0 | 0.0 | 0.0 | 0.0 |
| Urdu | $\Delta_\mu$ | −0.003 | −0.005 | −0.004 | −0.004 | −0.004 | −0.002 | −0.002 | −0.005 |
| | % | 0.11 | 0.22 | 0.2 | 0.26 | 0.24 | 0.14 | 0.13 | 0.37 |
| | $p$ | 0.0 | 0.0 | 0.0 | 0.0 | 0.0 | 0.02 | 0.217 | 0.001 |

Table 14: Significance tests for character $n$-gram language models. $\Delta_\mu$ is the mean absolute change in cross-entropy after normalization; % is the percentage change; and $p$ is the WS $p$-value.

The mean cross-entropy difference $\Delta_\mu$ is computed over all the $k$ folds as

$$\Delta_\mu = \frac{1}{k}\sum_{i=1}^{k}\left(H_i(T_i) - H_i(B_i)\right), \tag{1}$$

where the negative value of $\Delta_\mu$ indicates the decrease of character entropy $H$ of model $i$ compared to the baseline and hence constitutes an improvement.

As can be seen from the table, the $\Delta_\mu$ values are negative across the board, apart from the lowest $n$-gram orders ($n = 3$ and $n = 4$) for Sindhi. To determine whether these changes in cross-entropy are statistically significant, three types of tests (WS, MW and BM) were performed (see §5.1.1). All of the tests assess the null hypothesis that baseline and test configurations represent the same distribution. While we do not explicitly compute the correlation between the $p$-values for all the three tests, these tend to correlate with each other upon informal inspection. All significance test values for all languages are presented in Appendix B. Since the trends are largely the same, for ease of inspection we just show the WS $p$-value in Table 14, where the statistically significant degradation for Sindhi configuration corresponding to $n = 4$ is marked in red, and discuss the few disagreements in the Appendix.

**Word Models** We repeated all the experiments described in §5.1.3 above for the $n$-gram models trained on words for the orders 2, 3, 4 and 5. The details of the training and test splits are shown in Table 15. As mentioned above, the Kashmiri dataset is very small and this sparsity is only increased when considering word-sized tokens instead of characters (compare this with Table 13). This is reinforced by computing the word cross-entropies for Kashmiri models, shown in Figure 3, where the error band shows significantly higher variance (compared to the character models in Figure 2) across each $k$ splits for all the orders compared to other languages. Sindhi and Malay, which are the second and third smallest datasets, show a reasonably high variance as well, although it is significantly smaller than for Kashmiri. The plot for Kurdish (Sorani) indicates that the quality of the word models tends to degrate for this corpus beyond trigrams, possibly due to a relatively small size of the dataset.

| Language | Train $\mu$ | Train $\sigma$ | Test $\mu$ | Test $\sigma$ |
|---|---|---|---|---|
| azb | 9 704 485.4 | 7010.9 | 2 110 255.6 | 7010.9 |
| ckb | 10 462 576.0 | 9503.5 | 2 384 669.0 | 9503.5 |
| ks | 23 189.4 | 250.5 | 4164.6 | 250.5 |
| ms | 1 463 801.4 | 814.4 | 290 574.6 | 814.4 |
| pnb | 24 690 883.4 | 13 213.2 | 3 111 613.6 | 13 213.2 |
| sd | 4 680 624.1 | 662.0 | 74 124.0 | 662.0 |
| ug | 2 160 932.7 | 7224.1 | 515 917.3 | 7224.1 |
| ur | 37 234 659.5 | 23 873.1 | 9 235 049.5 | 23 873.1 |

Table 15: Word-level statistics for $k$-fold configurations and $m$ $n$-gram orders ($k = 100$, $m = 4$).

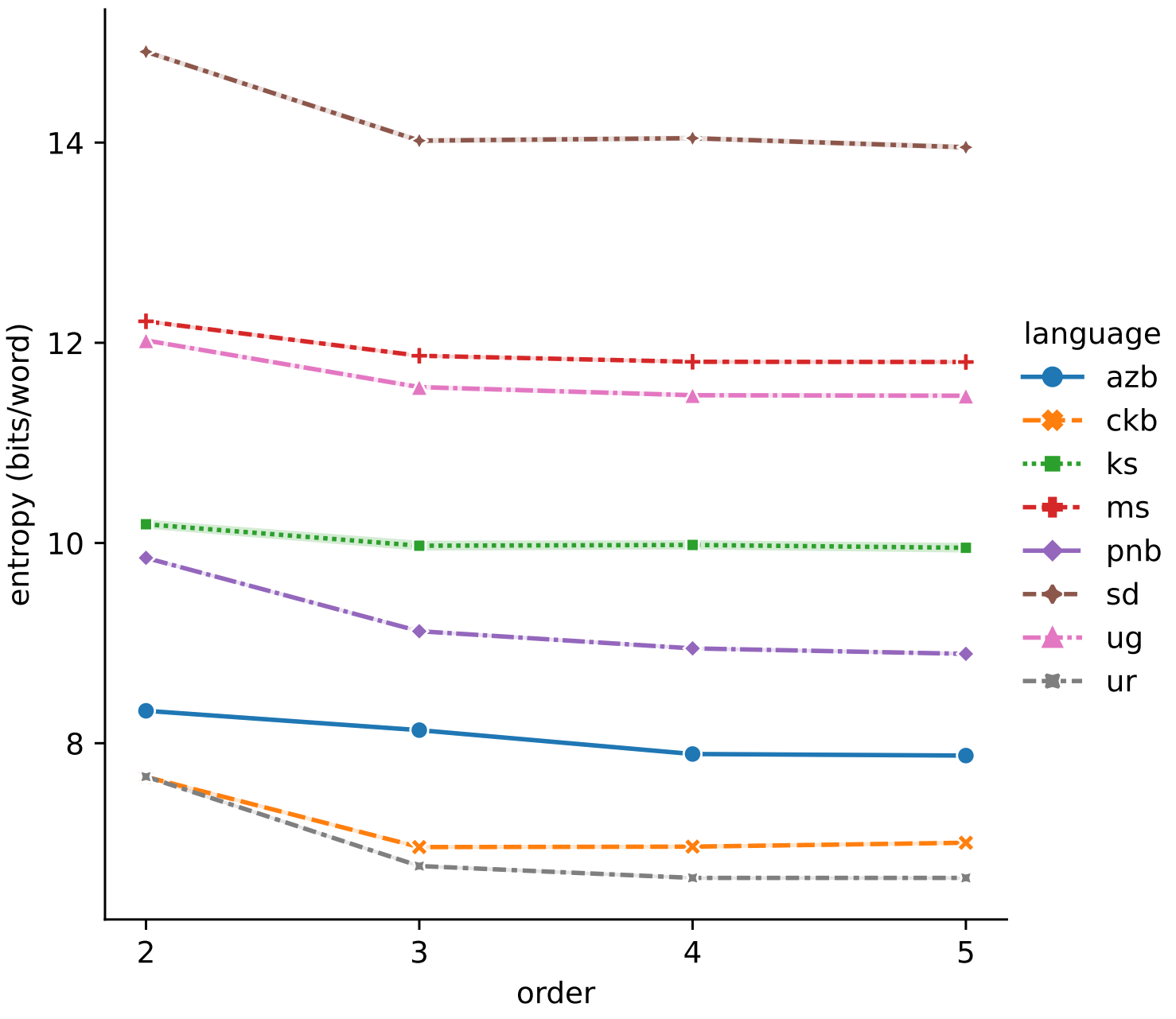

Figure 3: Average values of entropy (bits per word) over 100 runs vs. word $n$-gram orders.

| Language | Measure | *n*-gram order 2 | 3 | 4 | 5 |
|---|---|---|---|---|---|
| South Azerbaijani | $\Delta_\mu$ | −0.031 | −0.027 | −0.031 | −0.028 |
| | % | 0.374 | 0.332 | 0.397 | 0.358 |
| | $p$ | 0.0 | 0.0 | 0.0 | 0.0 |
| Kurdish (Sorani) | $\Delta_\mu$ | −0.031 | −0.034 | −0.034 | −0.035 |
| | % | 0.41 | 0.49 | 0.49 | 0.5 |
| | $p$ | 0.0 | 0.0 | 0.0 | 0.0 |
| Kashmiri | $\Delta_\mu$ | 0.034 | −0.039 | −0.022 | 0.017 |
| | % | −0.33 | 0.39 | 0.22 | −0.17 |
| | $p$ | 0.112 | 0.127 | 0.34 | 0.468 |
| Malay (Jawi) | $\Delta_\mu$ | −0.358 | −0.394 | −0.403 | −0.411 |
| | % | 2.935 | 3.319 | 3.411 | 3.479 |
| | $p$ | 0.0 | 0.0 | 0.0 | 0.0 |
| Punjabi (Shahmukhi) | $\Delta_\mu$ | −0.015 | −0.017 | −0.02 | −0.02 |
| | % | 0.15 | 0.19 | 0.23 | 0.23 |
| | $p$ | 0.0 | 0.0 | 0.0 | 0.0 |
| Sindhi | $\Delta_\mu$ | −0.159 | −0.169 | −0.177 | −0.184 |
| | % | 1.06 | 1.21 | 1.26 | 1.32 |
| | $p$ | 0.0 | 0.0 | 0.0 | 0.0 |
| Uyghur | $\Delta_\mu$ | −0.024 | −0.038 | −0.034 | −0.042 |
| | % | 0.2 | 0.332 | 0.294 | 0.367 |
| | $p$ | 0.0 | 0.0 | 0.0 | 0.0 |
| Urdu | $\Delta_\mu$ | −0.006 | −0.004 | −0.006 | −0.007 |
| | % | 0.08 | 0.06 | 0.1 | 0.1 |
| | $p$ | 0.0 | 0.001 | 0.0 | 0.0 |

Table 16: Significance tests for word *n*-gram language models. $\Delta_\mu$ is the mean absolute change in cross-entropy after normalization; % is the percentage change; and $p$ is the WS $p$-value.

Statistical significance tests were also performed for the word *n*-gram models constructed for $n \in [2, 3, 4, 5]$ from the $k = 100$ folds over original and normalized text. All values for all languages are presented in Appendix B, and key measures over all languages are shown in Table 16. Again, we just show the WS $p$-value in this summary table, but the values for all tests are presented and discussed in the Appendix. Kashmiri results are not significant for any of the orders, likely due to the very small size of the dataset. All other languages show statistically significant reductions in cross-entropy for all n-gram orders. Reductions are relatively small for Kurdish, Punjabi, Azerbaijani, Uyghur and Urdu, but more substantial for Sindhi and particularly Malay, which achieves up to 3.5% reduction in cross-entropy.

In sum, we have shown consistently small yet significant improvements in intrinsic language model quality through the use of these normalization methods.

## 5.2 Neural Machine Translation Experiments

This section describes the application of Perso-Arabic script normalization to machine translation (MT), which is arguably one of the oldest and most popular downstream NLP tasks (Hutchins, 1986). We selected a subset of languages described in §3 and designed a simple translation experiment, where for each language we build a model that translates that source language into English in two configurations: the model trained on the original source text and the model trained on the normalized source text. We hypothesize that if the normalization is "useful", it will result in a better model of the source language (by removing the extrinsic orthographic artifacts of Perso-Arabic resulting in systematic ambiguities) and, consequently, a better translation quality into English as measured by the objective evaluation metrics.

In what follows we introduce the parallel language corpora used for training and evaluating the models in §5.2.1, provide brief summary of the monolingual and multilingual model architectures used in §5.2.2, and discuss our results in §5.2.3. It is important to note that our aim here is not to produce a competitive MT system using current state-of-the-art (such as Wenzek et al., 2021; Xue et al., 2021, 2022), but rather to measure the effects of script normalization using reasonably advanced yet simple-to-train neural models.

| Languages | | Train | Development | Test |
|---|---|---|---|---|
| Kurdish | Corpora | XLEnt (ckb), TICO-19 (ckb), Wikimedia (ckb), OPUS-100 (kur), Tanzil (kur), Tatoeba (kur) | OPUS-100 (kur) | OPUS-100 (kur) |
| | Sizes | 256,909 | 2,000 | 2,000 |
| Sindhi | Corpora | CCMatrix, XLEnt, Tanzil, QED, Wikimedia | Ubuntu | CCMatrix (held-out) |
| | Sizes | 1,960,022 | 6,204 | 1,000 |
| Urdu | Corpora | OPUS-100, Joshua, Anuvaad | OPUS-100 | OPUS-100 |
| | Sizes | 798,574 | 2,736 | 2,000 |
| Uyghur | Corpora | XLEnt, Tanzil, Tatoeba, TED, OPUS-100 | OPUS-100 | OPUS-100 |
| | Sizes | 176,179 | 2,000 | 2,000 |

Table 17: Parallel corpora details for the four languages. The dataset sizes for each language correspond to the total number of parallel sentence pairs in all datasets for a particular language.

### 5.2.1 Parallel Corpora

In our experiments we construct individual models for translating from four languages into English: Kurdish, Sindhi, Urdu and Uyghur. These parallel corpora were collected using the MTData tool that automates the collection and preparation of machine translation datasets (Gowda et al., 2021).[45] Each language may have multiple datasets available from several sources, such as the OPUS collection that provides various machine translation corpora for many languages (Tiedemann, 2012).[46] The details of the corpora including all the datasets involved in training and testing the models for each language are shown in Table 17.

**Kurdish (including Sorani)** The amount of Sorani-specific parallel Sorani-English data available online is rather small, therefore we opted to also include general Kurdish-English parallel corpora (available under kur macrolanguage ISO 639-3 code) in our training data. The resulting model is in essence a multi-dialect and multi-script model for translating from Sorani (ckb), Kurmanji (kmr), or Northern Kurdish (that uses Latin script), and possibly other Kurdish dialects into English, but we make sure that we evaluate the model using test sentence pairs that include a substantial proportion of source sentences in Perso-Arabic script.

The training set includes the Sorani (ckb) part of XLEnt (El-Kishky et al., 2021), TICO-19 (Anastasopoulos et al., 2020),[47] and Wikimedia (Tiedemann, 2012),[48], as well as the general Kurdish (kur) training portions of OPUS-100 (Zhang et al., 2020),[49] the Tanzil corpus of religious texts,[50], and the Tatoeba dataset.[51] The development set consists of the development portion of OPUS-100 for general Kurdish. The test set is a general Kurdish test set of OPUS-100 dataset. In this set there are 329 source (Kurdish) sentences out of 2,000 which are in Perso-Arabic script.

**Sindhi** The Sindhi training data includes the Sindhi-English parallel data from the following datasets: XLEnt, Wikimedia, Tanzil, CCMatrix (Schwenk et al., 2021; Fan et al., 2021),[52] and QED (Abdelali et al., 2014).[53] For the development set we selected the Sindhi-English localization strings from the Ubuntu project.[54]. The test set consists of 1,000 sentence pairs withheld from the OPUS-100 training

[45] https://github.com/thammegowda/mtdata
[46] https://opus.nlpl.eu/
[47] https://opus.nlpl.eu/tico-19.php
[48] https://opus.nlpl.eu/wikimedia.php
[49] https://opus.nlpl.eu/opus-100.php
[50] https://opus.nlpl.eu/Tanzil.php (https://tanzil.net/)
[51] https://opus.nlpl.eu/Tatoeba.php (https://tatoeba.org/en/)
[52] https://opus.nlpl.eu/CCMatrix.php
[53] https://opus.nlpl.eu/QED.php
[54] https://opus.nlpl.eu/Ubuntu.php

set.

**Urdu** In the training data we include the Urdu-English sentence pairs from the Anuvaad corpus (Anuvaad, 2022), the parallel South Asian corpora from the Joshua statistical machine translation (SMT) toolkit (Post et al., 2012),[55] and the training set from OPUS-100. The development and test sets consist of the respective development and test Urdu-English partitions of the OPUS-100 dataset.

**Uyghur** The training data includes the Uyghur-English pairs from the following corpora: XLEnt, Tanzil, Tatoeba, TED (Reimers and Gurevych, 2020),[56] and the training partition of OPUS-100. The development and test sets consist of the respective development and test Uyghur-English partitions of the OPUS-100 dataset.

**Multilingual Configuration** In addition to constructing individual monolingual translation models we also experiment with a single multilingual model that provides many-to-English translation. Rather than using all the available data we constructed a corpus that is balanced in terms of per-language parallel sentence pairs: all of the training data is selected for Uyghur, which is our smallest dataset (see Table 17), and for the rest of the languages we selected the first 200,000 sentence pairs from the respective training sets. The resulting multilingual training set thus constructed consists of 776,179 sentence pairs. The test set for multilingual configuration consists of 7,000 sentence pairs that correspond to the whole test sets for the respective languages.

### 5.2.2 Models

Modern neural machine translation (NMT) models are an instance of neural sequence-to-sequence models, which have achieved impressive results in recent years (Stahlberg, 2020). In our experiments we use a variant of recurrent neural network (RNN) encoder-decoder bipartite architecture equipped with attention mechanism (Mnih et al., 2014; Bahdanau et al., 2015), where instead of RNN units, long short-term memory (LSTM) cells are used, which allows the network to learn the long sequences more efficiently (Hochreiter and Schmidhuber, 1997). The particular attention mechanism we use in the decoder is described in Luong et al. (2015).

We use two different model configurations. For languages with larger amount of data (Sindhi and Urdu, as shown in Table 17), the encoder component is bidirectional (Schuster and Paliwal, 1997), consisting of four stacked layers of 256 LSTM units each, while the decoder memory consists of 512 units. The configuration used for languages with smaller amounts of data, Kurdish and Uyghur, is mostly identical, but the encoder has two unidirectional LSTM layers. Both models correspond to vanilla configurations (`NMTMediumV1` and `NMTSmallV1`) provided by the OpenNMT-tf library (Klein et al., 2017) implemented in the TensorFlow framework (Abadi et al., 2016). Our models are word-based, with the 50,000 most frequent words used for source and target embedding vocabularies. The tokenization is performed in *aggressive* mode provided by the default OpenNMT tokenizer. The parallel sentence pairs where either the source or the target sentence is longer than 100 words are dropped from the training. Overall, the larger models `NMTMediumV1` have approximately 92M model parameters, while the smaller models `NMTSmallV1` have approximately 62M parameters. We used default hyperparameters provided by the OpenNMT configurations, apart from the training batch size which we set to 64 examples.

For the multilingual experiment, the size of the balanced dataset described in §5.2.1 roughly corresponds to the size of our Urdu corpus. Hence, similar to Urdu and Sindhi, we have chosen the `NMTMediumV1` configuration for our multilingual many-to-English model with the same hyperparameters as for the monolingual configurations.

| Epochs | Δ BLEU | Δ chrF2 | Δ TER |
|---|---|---|---|
| 1 | −3.216 | 30.460 | 0.665 |
| 2 | −10.719 | −5.340 | −0.883 |
| 3 | 2.249 | 8.010 | −3.095 |
| 4 | 30.132 | 14.786 | −2.996 |
| 5 | 28.440 | 17.847 | 1.985 |
| 6 | 20.165 | 16.023 | −2.022 |
| 7 | 17.866 | 12.143 | −6.269 |
| 8 | 31.357 | 18.202 | −4.183 |
| $\mu$ | 14.534 | 14.015 | −2.099 |

(a) Kurdish

| Epochs | Δ BLEU | Δ chrF2 | Δ TER |
|---|---|---|---|
| 1 | 11.239 | 2.264 | 4.059 |
| 2 | 6.358 | 2.351 | −3.655 |
| 3 | 3.536 | 0.028 | −2.644 |
| 4 | 7.950 | 3.158 | −2.161 |
| 5 | 2.024 | 1.542 | 4.414 |
| 6 | 1.075 | 0.652 | −0.503 |
| 7 | 3.384 | 3.173 | 1.117 |
| 8 | 1.254 | −0.298 | −2.795 |
| $\mu$ | 4.603 | 1.609 | −0.271 |

(b) Sindhi

| Epochs | Δ BLEU | Δ chrF2 | Δ TER |
|---|---|---|---|
| 1 | −2.063 | −2.960 | −2.657 |
| 2 | 10.752 | 3.428 | −1.981 |
| 3 | 6.686 | 1.269 | −2.686 |
| 4 | 8.741 | 4.109 | −0.764 |
| 5 | 4.781 | 1.423 | 0.121 |
| 6 | 3.657 | 0.610 | −1.366 |
| 7 | 4.634 | 1.606 | −2.558 |
| 8 | 3.367 | 1.726 | −0.441 |
| $\mu$ | 5.069 | 1.401 | −1.541 |

(c) Urdu

| Epochs | Δ BLEU | Δ chrF2 | Δ TER |
|---|---|---|---|
| 1 | −5.998 | −1.583 | 0.130 |
| 2 | 5.920 | 1.378 | 2.780 |
| 3 | 2.661 | 0.960 | 0.128 |
| 4 | 6.525 | 7.454 | −5.967 |
| 5 | 1.874 | 0.046 | 1.550 |
| 6 | −2.120 | −3.592 | −2.101 |
| 7 | −0.345 | 1.451 | 0.748 |
| 8 | 0.600 | 0.348 | 1.224 |
| $\mu$ | 1.140 | 0.808 | −0.188 |

(d) Uyghur

Table 18: Relative difference (%) between the performance of normalized and unnormalized models.

### 5.2.3 Results and Discussion

For each language two native language-to-English models were trained from unnormalized and normalized text for that language, respectively.[57] The details of the language-specific text partitions are provided in §5.2.1. Perso-Arabic script normalization was applied to the native language side of training, development and testing portions of the data, with the English side kept unchanged. Each model was trained for 8 epochs and at the end of each epoch model's performance was evaluated on the test set using the three MT metrics, each using default parameters such as casing and smoothing, provided by the SacreBLEU toolkit (Post, 2018):[58] the BiLingual Evaluation Understudy, or BLEU score (Papineni et al., 2002), the Character $n$-gram F-score, or chrF2 (Popović, 2016), and Translation Edit Rate, or TER (Snover et al., 2006). Higher BLEU and chrF2 scores indicate that the hypotheses better match the reference translations, whereas for TER lower scores indicate a better match.

The relative differences (in %) computed between the scores for the models build on normalized and unnormalized text for each language are shown in Table 18, with the positive values of Δ BLEU and Δ chrF2, and the negative values of Δ TER signifying relative improvement in performance of the normalized model over the unnormalized one at each training epoch. The last two highlighted rows in each table correspond to relative performance differences at the last epoch and the mean per-epoch difference $\mu$. The improvements are highlighted in green and the degradation in red.[59]

As can be seen from Table 18, the biggest gains over the baseline are obtained for the normalized Kurdish model over all the three MT metrics for both the last training epoch as well as the average per-epoch relative difference. The normalized Urdu model also displays improvements across the board. For

[55] https://github.com/joshua-decoder/indian-parallel-corpora

[56] https://opus.nlpl.eu/TED2020.php

[57] While we do not provide the statistics for the normalized NMT data in terms of number of training set lines, tokens and types changed by the normalization, we hypothesize that these ratios would be similar to the ones computed for statistical language modeling experiments using Wikipedia data presented in Table 12 in §5.1.

[58] https://github.com/mjpost/sacrebleu

[59] The absolute raw scores are also provided in Table 19.

| Systems | | PBS | | | PAR | | |
|---|---|---|---|---|---|---|---|
| | | BLEU ($\mu$ ± 95% CI) | chrF2 ($\mu$ ± 95% CI) | TER ($\mu$ ± 95% CI) | BLEU | chrF2 | TER |
| | $\mathcal{B}$ | 10.740 (10.731 ± 1.082) | 27.187 (27.193 ± 1.068) | 83.047 (83.070 ± 1.442) | 10.740 | 27.187 | 83.047 |
| Kurdish | $\mathcal{N}$ | 15.646 (15.617 ± 1.269) | 33.237 (33.220 ± 1.210) | 79.712 (79.742 ± 1.769) | 15.646 | 33.237 | 79.712 |
| | $p$ | 0.001 | 0.001 | 0.001 | 0.000 | 0.000 | 0.000 |
| | $\mathcal{B}$ | 15.362 (15.367 ± 0.675) | 39.223 (39.228 ± 0.707) | 74.960 (74.967 ± 1.034) | 15.362 | 39.223 | 74.960 |
| Sindhi | $\mathcal{N}$ | 15.557 (15.572 ± 0.717) | 39.106 (39.116 ± 0.719) | 72.921 (72.923 ± 0.991) | 15.557 | 39.106 | 72.921 |
| | $p$ | 0.148 | 0.199 | 0.001 | 0.392 | 0.575 | 0.000 |
| | $\mathcal{B}$ | 13.387 (13.393 ± 0.702) | 33.559 (33.562 ± 0.771) | 77.403 (77.402 ± 1.002) | 13.387 | 33.559 | 77.403 |
| Urdu | $\mathcal{N}$ | 13.854 (13.854 ± 0.709) | 34.148 (34.151 ± 0.751) | 77.064 (77.061 ± 1.066) | 13.854 | 34.148 | 77.064 |
| | $p$ | 0.043 | 0.009 | 0.167 | 0.091 | 0.010 | 0.442 |
| | $\mathcal{B}$ | 9.982 (9.938 ± 0.588) | 29.118 (29.117 ± 0.636) | 87.236 (87.207 ± 1.678) | 9.982 | 29.118 | 87.236 |
| Uyghur | $\mathcal{N}$ | 10.043 (10.017 ± 0.581) | 29.220 (29.225 ± 0.628) | 88.317 (88.352 ± 1.413) | 10.043 | 29.220 | 88.317 |
| | $p$ | 0.319 | 0.229 | 0.105 | 0.829 | 0.668 | 0.277 |

Table 19: Paired significance tests for the monolingual models obtained after the final training epoch: Paired Bootstrap Resampling (PBS) and Paired Approximate Randomization (PAR).

Sindhi, there is a relative degradation of 0.298% in chrF2 over the unnormalized baseline at the last training epoch, while for Uyghur there is a larger final-epoch degradation of 1.224% in TER. Apart from these two cases however, overall the mean per-epoch and final-epoch relative differences indicate potential improvements, although in the case of Uyghur these are small.

In order to ascertain whether the above relative differences are statistically significant we performed paired significance testing of the final-epoch systems using two algorithms provided by SacreBLEU: the Paired Bootstrap Resampling (Koehn, 2004) and the Paired Approximate Randomization (Riezler and Maxwell, 2005), denoted PBS and PAR, respectively. For PBS, the default parameter of 1,000 resampling trials was used. For PAR, the default value of 10,000 trials was used for randomization test. The results of both tests are shown in Table 19. For each language two systems are tested: the unnormalized baseline ($\mathcal{B}$), and the model built from the normalized text ($\mathcal{N}$). The systems are pairwise compared for sentences from the test set using the three MT metrics described above. The null hypotheses for both tests postulate that both $\mathcal{B}$ and $\mathcal{N}$ translations are generated by the same underlying process. For a given model $\mathcal{N}$ and the baseline $\mathcal{B}$, the $p$-value is roughly the probability of the absolute score difference ($\Delta$) or higher occurring due to chance, under the assumption that the null hypothesis is correct. Assuming a significance threshold of 0.05, the null hypothesis can be rejected for $p$-values $< 0.05$, which implies that both systems are different. For PBS, the actual system score, the bootstrap estimated true mean ($\mu$), and the 95% confidence interval (CI) are shown for each metric. For PAR, no true mean or confidence intervals are shown because the algorithm does not perform the resampling.

In Table 19 the statistically significant improvements are highlighted in green, while the cases where the systems appear to be equivalent are highlighted in light blue. We note that both small degradations in translation quality of Sindhi and Uyghur for the individual metrics observed in Table 18 turn out to be not statistically significant, as evidenced by the corresponding $p$-values. For Kurdish, the improvements are statistically significant across the board, while for Sindhi and Urdu the improvements are significant according to at least one MT metric and at least one significance test: both PBS and PAR agree on improvements in TER on Sindhi and in chrF2 on Urdu (where PBS also indicates significant improvement in BLEU). Interestingly, both tests indicate that normalization has no effect on Uyghur translation quality. We hypothesize that this may be due to several conflating factors. First, since this is the smallest dataset of all the languages in this experiment (see Table 17), there may not be enough data for training the model reliably. Furthermore, potential misalignment between Uyghur and English sentences in the training data may be adversely affecting the quality of the resulting models.

**Many-to-English Experiment** The goal of this experiment is to verify the hypothesis that the relative performance of Perso-Arabic script-normalized individual NMT systems, especially Uyghur, is improved by pooling the data from other available languages. To this end we trained a single many-to-English model described in §5.2.1 and §5.2.2. Similar to individual language-to-English experiments, we compare the performance of the NMT model built from normalized text against the baseline model constructed from unnormalized data.

| Epochs | Δ BLEU | Δ chrF2 | Δ TER |
|---|---|---|---|
| 1 | 31.456 | 12.001 | 5.988 |
| 2 | 35.790 | 18.781 | −4.837 |
| 3 | 28.537 | 14.952 | −2.812 |
| 4 | 25.456 | 13.398 | −5.170 |
| 5 | 22.387 | 9.781 | −4.486 |
| 6 | 14.101 | 7.564 | −6.108 |
| 7 | 15.732 | 8.336 | −4.087 |
| 8 | 10.666 | 5.702 | −6.660 |
| $\mu$ | 23.016 | 11.314 | −3.521 |

Table 20: Relative difference (%) between the performance of two (normalized and unnormalized) many-to-English models.

| Systems | | PBS | | | PAR | | |
|---|---|---|---|---|---|---|---|
| | | BLEU ($\mu$ ± 95% CI) | chrF2 ($\mu$ ± 95% CI) | TER ($\mu$ ± 95% CI) | BLEU | chrF2 | TER |
| Kurdish | $\mathcal{B}^m$ | 12.968 (12.891 ± 1.471) | 29.403 (29.412 ± 1.144) | 83.296 (83.300 ± 2.881) | 12.968 | 29.402 | 83.296 |
| | $\mathcal{N}^m$ | 18.496 (18.505 ± 1.511) | 34.773 (34.781 ± 1.334) | 73.373 (73.371 ± 1.820) | 18.496 | 34.773 | 73.373 |
| | $p$ | 0.001 | 0.001 | 0.001 | 0.000 | 0.000 | 0.000 |
| Sindhi | $\mathcal{B}^m$ | 14.602 (14.586 ± 0.690) | 37.889 (37.901 ± 0.649) | 76.918 (76.920 ± 1.105) | 14.602 | 37.889 | 76.918 |
| | $\mathcal{N}^m$ | 15.715 (15.727 ± 0.742) | 39.410 (39.425 ± 0.675) | 72.889 (72.895 ± 1.011) | 15.715 | 39.410 | 72.890 |
| | $p$ | 0.001 | 0.001 | 0.001 | 0.000 | 0.000 | 0.000 |
| Urdu | $\mathcal{B}^m$ | 11.226 (11.211 ± 0.621) | 30.516 (30.526 ± 0.677) | 84.246 (84.236 ± 1.651) | 11.226 | 30.516 | 84.246 |
| | $\mathcal{N}^m$ | 12.255 (12.247 ± 0.658) | 32.057 (32.057 ± 0.721) | 79.169 (79.185 ± 1.065) | 12.255 | 32.057 | 79.169 |
| | $p$ | 0.001 | 0.001 | 0.001 | 0.000 | 0.000 | 0.000 |
| Uyghur | $\mathcal{B}^m$ | 15.346 (15.361 ± 0.820) | 35.412 (35.419 ± 0.788) | 75.741 (75.731 ± 1.206) | 15.346 | 35.412 | 75.741 |
| | $\mathcal{N}^m$ | 17.031 (17.050 ± 0.881) | 37.377 (37.388 ± 0.877) | 71.677 (71.656 ± 1.182) | 17.031 | 37.377 | 71.677 |
| | $p$ | 0.001 | 0.001 | 0.001 | 0.000 | 0.000 | 0.000 |

Table 21: Paired significance tests for the multilingual model obtained after the final training epoch: Paired Bootstrap Resampling (PBS) and Paired Approximate Randomization (PAR).

Before proceeding two important points need to be noted. First, because our Perso-Arabic script normalization grammars are language-specific, the normalized version of the multilingual corpus described in §5.2.1 is constructed from the normalized corpora for the respective individual languages. Second, since our balanced multilingual corpus has roughly the same number of sentence pairs for each language, the amount of within-language data available for the "bigger" languages (Urdu and Sindhi) is significantly smaller in this experiment. Thus, when one compares NMT scores for these languages between a many-to-English system on one hand, and a particular monolingual language-to-English system on the other, the many-to-English scores may be worse. This is not an issue because, as mentioned above, the goal of this experiment is to investigate the relative improvements over the unnormalized baseline, rather than constructing an NMT system with the best possible absolute score.

The relative differences (in %) computed between the scores for the many-to-English models build from normalized and unnormalized text are shown in Table 20. It is worth noting that unlike the monolingual scores shown in Table 18 these scores are computed using the combined test data consisting of 7,000 sentence pairs from all the individual languages described in §5.2.1. As can be seen from Table 20, the normalized many-to-English model shows consistent improvements in all the metrics over the unnormalized baseline for all the epochs with the exception of 5.988% degradation in TER for the initial epoch.

We also performed PBS and PAR paired statistical significance tests for the many-to-English configuration comparing the performance of the multilingual normalized model (denoted $\mathcal{N}^m$) against its unnormalized counterpart ($\mathcal{B}^m$) on the test data for individual languages. The results of both tests are shown in Table 21 for each language, with the statistically significant improvements in individual metrics marked in green. Compared to significance tests for the monolingual systems in Table 19, the multilingual tests show more robust improvements across all languages and metrics. In particular, with respect to Uyghur these results confirm our hypothesis above that the original dataset is too small to reliably measure the effects of script normalization. This is rectified by using data from other languages. Additional supporting evidence comes from comparing between the absolute values of all the metrics for normalized and unnormalized monolingual and multilingual models for Uyghur and Kurdish shown in Table 19 and Table 21: multilingual configurations have higher absolute scores for these languages.

# 6 Conclusions

This paper provided a brief overview of various adaptations of the Perso-Arabic script for eight languages from diverse language families and the issues that result from representing these adaptations digitally. The particular emphasis of this study was on the visual ambiguities between Perso-Arabic characters represented in Unicode. We argued that the computational methods for visual disambiguation need to go beyond the standard language-agnostic techniques provided by the Unicode standard and take into account the specifics of a local writing system as well as multiple confounding factors that affect the patterns of its use. We presented two types of writing system-specific normalization methods. Similar to the standard Unicode normalization techniques, *visual* normalization preserves the visual invariance of the characters, while providing significantly broader coverage of normalization cases peculiar to the orthography in question. The second type is *reading* normalization, which provides character transformations that violate visual invariance (e.g., by modifying the number of *i'jām* dots on the base shape of a character), yet are required to make the input conform to the local orthography. The distinction between the two types hinges on the visual invariance criterion, which is helpful in deciding when and if to apply either type of technique.[60] Perso-Arabic script normalization techniques are crucial for cybersecurity, but the focus of this paper is on their application to natural language processing. We performed experiments in statistical language modeling and neural machine translation that demonstrated the positive impact of script normalization on the performance of the resulting models.

[60] In some applications it may be necessary to preserve the visual fidelity of the input, hence the application of reading normalization may not be desirable.

This study describes work that is still in early stages. While there is a wealth of literature on the eight languages described in this paper and some additional languages currently covered by our methods, the majority of Perso-Arabic writing systems are used for lower-resource languages that are either scarcely documented or have very little online data, which is needed to provide evidence for required normalization. Furthermore, significant research towards a formal description of Perso-Arabic script typology is still required.

## Acknowledgements

The authors would like to thank Christo Kirov for many useful comments on an earlier draft of this paper, as well as Lawrence Wolf-Sonkin, Işın Demirşahin and Anna Katanova for informative feedback and assistance with the various stages of this project. We also thank Aso Mahmudi for his invaluable feedback on the modern orthography of Central Kurdish.

## References

Muzaffar Aazim, Kamal Mansour, and Roozbeh Pournader. 2009. Proposal to add two Kashmiri characters and one annotation to the Arabic block. Technical Report L2/09-176, Unicode Consortium.

Martín Abadi, Ashish Agarwal, Paul Barham, Eugene Brevdo, Zhifeng Chen, Craig Citro, Gregory S. Corrado, Andy Davis, Jeffrey Dean, Matthieu Devin, Sanjay Ghemawat, Ian J. Goodfellow, Andrew Harp, Geoffrey Irving, Michael Isard, Yangqing Jia, Rafal Józefowicz, Lukasz Kaiser, Manjunath Kudlur, Josh Levenberg, Dan Mané, Rajat Monga, Sherry Moore, Derek Gordon Murray, Chris Olah, Mike Schuster, Jonathon Shlens, Benoit Steiner, Ilya Sutskever, Kunal Talwar, Paul A. Tucker, Vincent Vanhoucke, Vijay Vasudevan, Fernanda B. Viégas, Oriol Vinyals, Pete Warden, Martin Wattenberg, Martin Wicke, Yuan Yu, and Xiaoqiang Zheng. 2016. TensorFlow: Large-Scale Machine Learning on Heterogeneous Distributed Systems. *CoRR*, abs/1603.04467.

Ahmed Abdelali, Francisco Guzman, Hassan Sajjad, and Stephan Vogel. 2014. The AMARA corpus: Building parallel language resources for the educational domain. In *Proceedings of the Ninth International Conference on Language Resources and Evaluation (LREC'14)*, pages 1856–1862, Reykjavik, Iceland. European Language Resources Association (ELRA).

Farhanah Abdullah, Asyraf Hj Ab Rahman, Abdul Manan Ali, Firdaus Khairi Abdul Kadir, and Fadzli Adam. 2020. Jawi script and the Malay society: Historical background and development. *International Journal of Management (IJM)*, 11(7):68–78.

Humza Ahmad and Laszlo Erdodi. 2021. Overview of phishing landscape and homographs in Arabic domain names. *Security and Privacy*, 4(4):1–14.

Sina Ahmadi. 2019. A rule-based Kurdish text transliteration system. *ACM Transactions on Asian and Low-Resource Language Information Processing (TALLIP)*, 18(2):1–8.

Antonios Anastasopoulos, Alessandro Cattelan, Zi-Yi Dou, Marcello Federico, Christian Federmann, Dmitriy Genzel, Franscisco Guzmán, Junjie Hu, Macduff Hughes, Philipp Koehn, Rosie Lazar, Will Lewis, Graham Neubig, Mengmeng Niu, Alp Öktem, Eric Paquin, Grace Tang, and Sylwia Tur. 2020. TICO-19: the translation initiative for COvid-19. In *Proceedings of the 1st Workshop on NLP for COVID-19 (Part 2) at EMNLP 2020*, Online. Association for Computational Linguistics.

Team Anuvaad. 2022. Project anuvaad: Parallel corpus. Accessed: June, 2022; `https://anuvaad.org/`.

Gulayhan Aqtay. 2020. The new Kazakh alphabet based on Latin script. In Gulayhan Aqtay and Cem Erdem, editors, *Language and Society in Kazakhstan: The Kazakh Context*, volume 6 of *Turkic Studies*, pages 23–36. Wydawnictwo Naukowe, Adam Mickiewicz University, Poznań, Poland.

Howard Aronson. 1996. Yiddish. In Peter Daniels and William Bright, editors, *The World's Writing Systems*, chapter 61, pages 735–742. Oxford University Press, Oxford.

Dmitry Bahdanau, Kyunghyun Cho, and Yoshua Bengio. 2015. Neural machine translation by jointly learning to align and translate. In *Proceedings of 3rd International Conference on Learning Representations (ICLR)*, San Diego, CA.

Elena Bashir. 2015. The Brahui language: Recovering the past, documenting the present, and pondering the future. In *Proceedings of International Conference on Brahui Language and Culture*, pages 1–27, Islamabad, Pakistan. Brahui Academy, Baluchistan, Pakistan.

Elena Bashir and Thomas J. Conners. 2019. *A Descriptive Grammar of Hindko, Panjabi, and Saraiki*, volume 4 of *Mouton-CASL Grammar Series (MCASL)*. De Gruyter Mouton.

Elena Bashir, Sarmad Hussain, and Deborah Anderson. 2006. Proposal for characters for Khowar, Torwali, and Burushaski. Technical Report L2-06/149, Unicode Consortium.

Thomas Bauer. 1996. Arabic writing. In Peter Daniels and William Bright, editors, *The World's Writing Systems*, chapter 50, pages 559–563. Oxford University Press, Oxford.

Clifford Edmund Bosworth. 2011. Arrān. In Ehsan Yarshater, editor, *Encyclopædia Iranica*, volume II/5, pages 520–522. Brill, Leiden, The Netherlands. Online.

Sami Boudelaa, Manuel Perea, and Manuel Carreiras. 2020. Matrices of the frequency and similarity of Arabic letters and allographs. *Behavior Research Methods*, 52(5):1893–1905.

Hamit Bozarslan, Cengiz Gunes, and Veli Yadirgi. 2021. Kurdish language. In *The Cambridge History of the Kurds*, pages 601–684. Cambridge University Press, Cambridge, UK. Part V.

Vladimir I. Braginsky. 1975. Some remarks on the structure of the "Sya'ir Perahu" by Hamzah Fansuri. *Bijdragen tot de Taal-, Land- en Volkenkunde / Journal of the Humanities and Social Sciences of Southeast Asia and Oceania*, 131(4):407–426.

William Bright. 1999. A matter of typology: Alphasyllabaries and abugidas. *Written Language & Literacy*, 2(1):45–55.

Michael C. Brose. 2017. The medieval Uyghurs of the 8th through 14th centuries. In *Oxford Research Encyclopedia of Asian History*, pages 1–20. Oxford University Press. Online.

Edgar Brunner and Ullrich Munzel. 2000. The nonparametric Behrens-Fisher problem: Asymptotic theory and small-sample approximation. *Biometrical Journal: Journal of Mathematical Methods in Biosciences*, 42(1):17–25.

Harun Buljina. 2019. *Empire, Nation, and the Islamic World: Bosnian Muslim Reformists between the Habsburg and Ottoman Empires, 1901–1914*. Ph.D. thesis, Columbia University, New York.

George L. Campbell. 1994. Kurdish. In George L. Campbell, editor, *Concise Compendium of the World's Languages*, chapter 5, pages 288–292. Routledge, London.

Rehmat Aziz Khan Chitrali. 2020a. Proposal to include Indus Kohistani language alphabets. Technical Report L2/20-157, Unicode Consortium.

Rehmat Aziz Khan Chitrali. 2020b. Proposal to include Kalasha language alphabets. Technical Report L2/20-091, Unicode Consortium.

Paolo Coluzzi. 2020. Jawi, an endangered orthography in the Malaysian linguistic landscape. *International Journal of Multilingualism*, pages 1–17.

Alexis Conneau, Kartikay Khandelwal, Naman Goyal, Vishrav Chaudhary, Guillaume Wenzek, Francisco Guzmán, Edouard Grave, Myle Ott, Luke Zettlemoyer, and Veselin Stoyanov. 2020. Unsupervised cross-lingual representation learning at scale. In *Proceedings of the 58th Annual Meeting of the Association for Computational Linguistics*, pages 8440–8451, Online. Association for Computational Linguistics.

Florian Coulmas. 1999. *The Blackwell Encyclopedia of Writing Systems*. John Wiley & Sons, Oxford, UK.

Peter T. Daniels. 2013. The Arabic writing system. In Jonathan Owens, editor, *The Oxford Handbook of Arabic Linguistics*, chapter 18, pages 422–431. Oxford University Press, Oxford, UK.

Peter T. Daniels. 2014. The type and spread of Arabic script. In Meikal Mumin and Kees Versteegh, editors, *The Arabic Script in Africa*, volume 71 of *Studies in Semitic Languages and Linguistics*, pages 25–39. Brill.

A. K. Datta. 1984. A generalized formal approach for description and analysis of major Indian scripts. *IETE Journal of Research*, 30(6):155–161.

Derek Davis. 2015. Premchand plays chess. *Journal of the Royal Asiatic Society*, 25(2):269–300.

DBP. 2006. *Daftar kata bahasa Melayu: Rumi-Sebutan-Jawi [Malay Pronunciation Dictionary (Rumi-Jawi)]*, 5th edition. Dewan Bahasa Pustaka (DBP) [Institute of Language and Literature], Kuala Lumpur, Malaysia.

Nuria de Castilla. 2019. Uses and written practices in Aljamiado manuscripts. In Dmitry Bondarev, Alessandro Gori, and Lameen Souag, editors, *Creating Standards: Interactions with Arabic script in 12 manuscript cultures*, volume 16 of *Studies in Manuscript Cultures*. De Gruyter.

Michael Dillon. 2009. *Xinjiang — China's Muslim Far Northwest*. Durham East Asia Series. Routledge, London and New York.

Hugh Dow. 1976. A note on the Sindhi alphabet. *Asian Affairs*, 7(1):54–56.

Arienne M. Dwyer. 2005. *The Xinjiang conflict: Uyghur identity, language policy, and political discourse*, volume 15 of *Policy Studies*. East-West Center Washington, Washington, D.C.

Ahmed El-Kishky, Adithya Renduchintala, James Cross, Francisco Guzmán, and Philipp Koehn. 2021. XLEnt: Mining a large cross-lingual entity dataset with lexical-semantic-phonetic word alignment. In *Proceedings of the 2021 Conference on Empirical Methods in Natural Language Processing*, pages 10424–10430, Online and Punta Cana, Dominican Republic. Association for Computational Linguistics.

Yahia Elsayed and Ahmed Shosha. 2018. Large scale detection of IDN domain name masquerading. In *Proceedings of 2018 APWG Symposium on Electronic Crime Research (eCrime)*, pages 1–11, San Diego, CA. IEEE.

Kyumars Sheykh Esmaili, Donya Eliassi, Shahin Salavati, Purya Aliabadi, Asrin Mohammadi, Somayeh Yosefi, and Shownem Hakimi. 2013. Building a test collection for Sorani Kurdish. In *Proceedings of ACS International Conference on Computer Systems and Applications (AICCSA)*, pages 1–7, Ifrane, Morocco. IEEE.

Lorna Priest Evans and Andy Warren-Rothlin. 2018. Proposal to encode additional Arabic script characters for Hausa to the UCS. Technical Report L2/18-094, Unicode Consortium.

Zohar Eviatar and Raphiq Ibrahim. 2014. Why is it hard to read Arabic? In Elinor Saiegh-Haddad and R. Malatesha Joshi, editors, *Handbook of Arabic Literacy: Insights and Perspectives*, volume 9 of *Literacy Studies*, pages 77–96. Springer.

Angela Fan, Shruti Bhosale, Holger Schwenk, Zhiyi Ma, Ahmed El-Kishky, Siddharth Goyal, Mandeep Baines, Onur Celebi, Guillaume Wenzek, Vishrav Chaudhary, et al. 2021. Beyond English-centric multilingual machine translation. *Journal of Machine Learning Research*, 22(107):1–48.

Liudmila L Fedorova. 2012. The development of structural characteristics of Brahmi script in derivative writing systems. *Written Language & Literacy*, 15(1):1–25.

Kyle Gorman. 2016. Pynini: A Python library for weighted finite-state grammar compilation. In *Proceedings of the SIGFSM Workshop on Statistical NLP and Weighted Automata*, pages 75–80, Berlin, Germany. Association for Computational Linguistics.

Kyle Gorman and Richard Sproat. 2021. *Finite-State Text Processing*. Number 50 in Synthesis Lectures on Human Language Technologies. Morgan & Claypool Publishers.

Thamme Gowda, Zhao Zhang, Chris Mattmann, and Jonathan May. 2021. Many-to-English machine translation tools, data, and pretrained models. In *Proceedings of the 59th Annual Meeting of the Association for Computational Linguistics and the 11th International Joint Conference on Natural Language Processing: System Demonstrations*, pages 306–316, Online. Association for Computational Linguistics.

Jagtar Singh Grewal. 2004. Historical geography of the Punjab. *Journal of Punjab Studies*, 11(1):1–18. Center of Sikh and Punjab Studies, UC Santa Barbara.

Beatrice Gruendler. 1993. *The development of the Arabic scripts: From the Nabatean era to the first Islamic century according to dated texts*, volume 43 of *Harvard Semitic Studies*. Scholars Press, Atlanta, Georgia.

Alexander Gutkin, Cibu Johny, Raiomond Doctor, Brian Roark, and Richard Sproat. 2022a. Beyond Arabic: Software for Perso-Arabic script manipulation. In *Proceedings of the 7th Arabic Natural Language Processing Workshop (WANLP2022)*, Abu Dhabi, United Arab Emirates. Association for Computational Linguistics. To appear.

Alexander Gutkin, Cibu Johny, Raiomond Doctor, Lawrence Wolf-Sonkin, and Brian Roark. 2022b. Extensions to Brahmic script processing within the Nisaba library: new scripts, languages and utilities. In *Proceedings of the Language Resources and Evaluation Conference*, pages 6450–6460, Marseille, France. European Language Resources Association.

Geoffrey Haig. 2018. The Iranian languages of northern Iraq. In Geoffrey Haig and Geoffrey Khan, editors, *The Languages and Linguistics of Western Asia: An Areal Perspective*, volume 6 of *The World of Linguistics (WOL)*, chapter 3.3, pages 267–304. Walter de Gruyter, Berlin, Munich & Boston.

Bahargül Hamut and Agnieszka Joniak-Lüthi. 2015. The language choices and script debates among the Uyghur in Xinjiang Uyghur Autonomous Region, China. *Linguistik Online*, 70(1):111–124.

Yannis Haralambous. 2021. Breaking Arabic: the creative inventiveness of Uyghur script reforms. *Design Regression*.

Lynley Hatcher. 2008. Script change in Azerbaijan: Acts of identity. *International Journal of the Sociology of Language*, 192:105–116.

Kenneth Heafield. 2011. KenLM: Faster and smaller language model queries. In *Proceedings of the Sixth Workshop on Statistical Machine Translation*, pages 187–197, Edinburgh, Scotland. Association for Computational Linguistics.

Kenneth Heafield, Ivan Pouzyrevsky, Jonathan H. Clark, and Philipp Koehn. 2013. Scalable modified Kneser-Ney language model estimation. In *Proceedings of the 51st Annual Meeting of the Association for Computational Linguistics (Volume 2: Short Papers)*, pages 690–696, Sofia, Bulgaria. Association for Computational Linguistics.

Ehab W. Hermena and Erik D. Reichle. 2020. Insights from the study of Arabic reading. *Language and Linguistics Compass*, 14(10):1–26.

Sepp Hochreiter and Jürgen Schmidhuber. 1997. Long short-term memory. *Neural Computation*, 9(8):1735–1780.

Sarmad Hussain, Ahmed Bakhat, Nabil Benamar, Meikal Mumin, and Inam Ullah. 2016. Enabling multilingual domain names: addressing the challenges of the Arabic script top-level domains. *Journal of Cyber Policy*, 1(1):107–129.

William John Hutchins. 1986. *Machine Translation: Past, Present, Future*. Ellis Horwood Series in Engineering Science. Ellis Horwood Chichester, Chichester (West Sussex), UK.

ICANN. 2011. Arabic case study team: Arabic case study team issues report. Internationalized Domain Names (IDN) Variant Issues project, Internet Corporation for Assigned Names and Numbers (ICANN).

ICANN. 2015. Task force on Arabic script IDN (TF-AIDN): Proposal for Arabic script Root Zone LGR. ICANN Internationalized Domain Names (IDN) program: Proposal documentation, Internet Corporation for Assigned Names and Numbers (ICANN). Version 2.7.

Arvind Iyengar. 2018. Variation in Perso-Arabic and Devanāgarī Sindhī orthographies: An overview. *Written Language & Literacy*, 21(2):169–197.

Carina Jahani and Agnes Korn. 2013. Balochi. In Gernot Windfuhr, editor, *The Iranian Languages*, Routledge Language Family Series, pages 710–768. Routledge.

Lars Johanson and Éva Ágnes Csató. 2021. *The Turkic Languages*, 2nd edition. Routledge Language Family Series. Routledge.

Cibu Johny, Lawrence Wolf-Sonkin, Alexander Gutkin, and Brian Roark. 2021. Finite-state script normalization and processing utilities: The Nisaba Brahmic library. In *Proceedings of the 16th Conference of the European Chapter of the Association for Computational Linguistics: System Demonstrations*, pages 14–23, Online. Association for Computational Linguistics.

Braj B. Kachru. 2016. Kashmiri and other Dardic languages. In Murray B. Emeneau and Charles A. Fergusson, editors, *Linguistics in South Asia*, volume 5 of *Current Trends in Linguistics*, pages 284–306. De Gruyter Mouton.

Raj Kachru. 2021. *Kashmiri Proverbs*. Notion Press Media, Chennai, Tamil Nadu, India.

Andreas Kaplony. 2008. What are those few dots for? Thoughts on the orthography of the Qurra Papyri (709–710), the Khurasan Parchments (755–777) and the inscription of the Jerusalem Dome of the Rock (692). *Arabica*, 55(Fasc. 1):91–112.

Alan S. Kaye. 1996. Adaptations of Arabic script. In Peter Daniels and William Bright, editors, *The World's Writing Systems*, chapter 62, pages 743–762. Oxford University Press, Oxford.

Hewa Salam Khalid. 2015. Kurdish dialect continuum, as a standardization solution. *International Journal of Kurdish Studies*, 1(1):27–39.

Nasha Bin Rodziadi Khaw. 2015. Study and analysis of the Proto-Shāradā and Shāradā inscriptions in the Lahore Museum (Pakistan). *Gandharan Studies*, 9:87–114.

Guillaume Klein, Yoon Kim, Yuntian Deng, Jean Senellart, and Alexander Rush. 2017. OpenNMT: Open-source toolkit for neural machine translation. In *Proceedings of ACL 2017, System Demonstrations*, pages 67–72, Vancouver, Canada. Association for Computational Linguistics.

Philipp Koehn. 2004. Statistical significance tests for machine translation evaluation. In *Proceedings of the 2004 Conference on Empirical Methods in Natural Language Processing*, pages 388–395, Barcelona, Spain. Association for Computational Linguistics.

Ernst Kotzé. 2012. Arabic Afrikaans – early standardisation of Afrikaans orthography: A discussion of the Afrikaans of the Cape Muslims by Achmat Davids. *Southern African Linguistics and Applied Language Studies*, 30(3):413–427.

Omkar Nath Koul. 2006. *A Dictionary of Kashmiri Proverbs*. Educa Books.

Omkar Nath Koul and Kashi Wali. 2015. *Kashmiri*, volume 03 of *LINCOM Grammar Handbooks (LGH)*. LINCOM Academic Publishers, Munich, Germany.

E. Ulrich Kratz. 2002. Jawi spelling and orthography: a brief review. *Indonesia and the Malay World*, 30(86):21–26.

Dennis Kurzon. 2013. Diacritics and the Perso-Arabic script. *Writing Systems Research*, 5(2):234–243.

Ali Murad Lajwani and Abdul Jaleel Mirjat. 2021. The mystical philosophy of Shah Abdul Latif Bhittai: A study of Shah-Jo-Risalo. *Al-Hikmat: A Journal of Philosophy*, 41:61–71.

K. Lekhwani and B. Lekhwani. 2014. *Sindhi Word Forms*. Chattisgarh Sindhi Sahitya Sansthan (Akademi) [Chattisgarh Sindhi Academy of Letters], Raipur, India.

David Lelyveld. 1994. *Zuban-e Urdu-e Mu'alla* and the idol of linguistic origins. *Annual of Urdu Studies*, 9.

Henrik Liljegren. 2016. *A grammar of Palula*, volume 8 of *Studies in Diversity Linguistics*. Language Science Press, Berlin, Germany.

Henrik Liljegren. 2018. Supporting and sustaining language vitality in Northern Pakistan. In Leanne Hinton, Leena Huss, and Gerald Roche, editors, *The Routledge Handbook of Language Revitalization*, pages 427–437. Routledge.

Thang Luong, Hieu Pham, and Christopher D. Manning. 2015. Effective approaches to attention-based neural machine translation. In *Proceedings of the 2015 Conference on Empirical Methods in Natural Language Processing*, pages 1412–1421, Lisbon, Portugal. Association for Computational Linguistics.

Henry B. Mann and Donald R. Whitney. 1947. On a test of whether one of two random variables is stochastically larger than the other. *The Annals of Mathematical Statistics*, 18(1):50–60.

Thomas Milo. 2002. Authentic Arabic: A case study. right-to-left font structure, font design, and typography. *Manuscripta Orientalia*, 8(1):49–61.

Volodymyr Mnih, Nicolas Heess, Alex Graves, and Koray Kavukcuoglu. 2014. Recurrent models of visual attention. In *Proceedings of Neural Information Processing Systems (NIPS)*, pages 2204–2212, Montreal, Canada.

Sayed Mohamed. 1968. *The Value of Dakhni Language and Literature (Special Lectures)*. University of Mysore, Mysore, India.

Mehryar Mohri. 1996. On some applications of finite-state automata theory to natural language processing. *Natural Language Engineering*, 2(1):61–80.

Mehryar Mohri. 2009. Weighted automata algorithms. In Manfred Droste, Werner Kuich, and Heiko Vogler, editors, *Handbook of Weighted Automata*, Monographs in Theoretical Computer Science, pages 213–254. Springer.

Payam Ghaffarvand Mokari and Stefan Werner. 2017. Azerbaijani. *Journal of the International Phonetic Association*, 47(2):207–212.

MS. 2012. Information technology — Jawi coded character set for information exchange. Malaysian Standard MS 2443:2012, Department of Standards Malaysia, Ministry of International Trade and Industry (MITI).

Benjamin Muller, Antonios Anastasopoulos, Benoît Sagot, and Djamé Seddah. 2021. When being unseen from mBERT is just the beginning: Handling new languages with multilingual language models. In *Proceedings of the 2021 Conference of the North American Chapter of the Association for Computational Linguistics: Human Language Technologies*, pages 448–462, Online. Association for Computational Linguistics.

Meikal Mumin. 2014. The Arabic script in Africa: Understudied literacy. In Meikal Mumin and Kees Versteegh, editors, *The Arabic Script in Africa*, volume 71 of *Studies in Semitic Languages and Linguistics*, pages 41–76. Brill, Leiden, The Netherlands.

C. Mohammed Naim. 1971. Arabic orthography and some non-Semitic languages. In Girdhari L. Tikku, editor, *Islam and its Cultural Divergence: Studies in Honor of Gustave E. von Grunebaum*, pages 113–144. University of Illinois Press, Urbana, IL.

Titus Nemeth. 2017. *Arabic Type-Making in the Machine Age: The Influence of Technology on the Form of Arabic Type, 1908–1993*, volume 14 of *Islamic Manuscripts and Books*. Brill, Leiden, The Netherlands.

Fallou Ngom and Mustapha H. Kurfi. 2017. ʿAjamization of Islam in Africa. *Islamic Africa*, 8(1-2):1–12.

Kishore Papineni, Salim Roukos, Todd Ward, and Wei-Jing Zhu. 2002. Bleu: a method for automatic evaluation of machine translation. In *Proceedings of the 40th Annual Meeting of the Association for Computational Linguistics*, pages 311–318, Philadelphia, Pennsylvania, USA. Association for Computational Linguistics.

Behrooz Parhami. 2020. Computers and challenges of writing in Persian: Explorations at the intersection of culture and technology. *Visible Language*, 54(1-2):187–223.

Neil Patel, Charles Riley, and Jesus MacLean. 2019. Proposal to add Arabic letter JEEM WITH THREE DOTS ABOVE and JEEM WITH THREE DOTS BELOW. Technical Report L2/19-118, Unicode Consortium.

Edoardo Maria Ponti, Ivan Vulić, Ryan Cotterell, Roi Reichart, and Anna Korhonen. 2019. Towards zero-shot language modeling. In *Proceedings of the 2019 Conference on Empirical Methods in Natural Language Processing and the 9th International Joint Conference on Natural Language Processing (EMNLP-IJCNLP)*, pages 2900–2910, Hong Kong, China. Association for Computational Linguistics.

Maja Popović. 2016. chrF deconstructed: beta parameters and n-gram weights. In *Proceedings of the First Conference on Machine Translation: Volume 2, Shared Task Papers*, pages 499–504, Berlin, Germany. Association for Computational Linguistics.

Matt Post. 2018. A call for clarity in reporting BLEU scores. In *Proceedings of the Third Conference on Machine Translation: Research Papers*, pages 186–191, Brussels, Belgium. Association for Computational Linguistics.

Matt Post, Chris Callison-Burch, and Miles Osborne. 2012. Constructing parallel corpora for six Indian languages via crowdsourcing. In *Proceedings of the Seventh Workshop on Statistical Machine Translation*, pages 401–409, Montréal, Canada. Association for Computational Linguistics.

Roozbeh Pournader. 2010. Of *hamza* and other *harakat*. Technical Report L2/10-455, Unicode Consortium.

D. J. Prentice. 1990. Malay (Indonesian and Malaysian). In Bernard Comrie, editor, *The Major Languages of East and South-East Asia*, chapter 10, pages 185–207. Routledge, London, UK.

Tahera Qutbuddin. 2007. Arabic in India: A survey and classification of its uses, compared with Persian. *Journal of the American Oriental Society*, 127(3):315–338.

Nils Reimers and Iryna Gurevych. 2020. Making monolingual sentence embeddings multilingual using knowledge distillation. In *Proceedings of the 2020 Conference on Empirical Methods in Natural Language Processing (EMNLP)*, pages 4512–4525, Online. Association for Computational Linguistics.

Ronit Ricci. 2011. *Islam translated: Literature, conversion, and the Arabic cosmopolis of South and Southeast Asia*. South Asia Across the Disciplines. University of Chicago Press, Chicago, IL, USA.

Stefan Riezler and John T. Maxwell. 2005. On some pitfalls in automatic evaluation and significance testing for MT. In *Proceedings of the ACL Workshop on Intrinsic and Extrinsic Evaluation Measures for Machine Translation and/or Summarization*, pages 57–64, Ann Arbor, Michigan. Association for Computational Linguistics.

Ronald Rosenfeld. 2000. Two decades of statistical language modeling: where do we go from here? *Proceedings of the IEEE*, 88(8):1270–1278.

Edward C. Sachau. 1910. *Alberuni's India: An Account of the Religion, Philosophy, Literature, Geography, Chronology, Astronomy, Customs, Laws and Astrology of India about A.D. 1030*. Kegan Paul, Trench, Trübner & Co., London.

Franklin E. Satterthwaite. 1946. An approximate distribution of estimates of variance components. *Biometrics Bulletin*, 2(6):110–114.

Ruth Laila Schmidt. 2007. Urdu. In George Cardona and Dhanesh Jain, editors, *The Indo-Aryan Languages*, Routledge Language Family Series, pages 315–385. Routledge.

Mike Schuster and Kuldip K. Paliwal. 1997. Bidirectional recurrent neural networks. *IEEE Transactions on Signal Processing*, 45(11):2673–2681.

Holger Schwenk, Guillaume Wenzek, Sergey Edunov, Edouard Grave, Armand Joulin, and Angela Fan. 2021. CCMatrix: Mining billions of high-quality parallel sentences on the web. In *Proceedings of the 59th Annual Meeting of the Association for Computational Linguistics and the 11th International Joint Conference on Natural Language Processing (Volume 1: Long Papers)*, pages 6490–6500, Online. Association for Computational Linguistics.

Rikza F. Sh. 2022. Proposal to encode three Quranic Arabic characters. Technical Report L2-22/153, Unicode Consortium.

Surinder Singh and Ishwar Dayal Gaur. 2009. *Sufism in Punjab: Mystics, Literature, and Shrines*. Aakar Books, Delhi, India.

Matthew Snover, Bonnie Dorr, Rich Schwartz, Linnea Micciulla, and John Makhoul. 2006. A study of translation edit rate with targeted human annotation. In *Proceedings of the 7th Conference of the Association for Machine Translation in the Americas: Technical Papers*, pages 223–231, Cambridge, Massachusetts, USA. Association for Machine Translation in the Americas.

Richard Sproat. 2003. A formal computational analysis of Indic scripts. In *In International Symposium on Indic Scripts: Past and Future*, Tokyo, Japan.

Felix Stahlberg. 2020. Neural machine translation: A review. *Journal of Artificial Intelligence Research*, 69:343–418.

Mikko Suutarinen. 2013. Arabic script among China's muslims: A Dongxiang folk story. In Tiina Hyytiäinen, Lotta Jalava, Janne Saarikivi, and Erika Sandman, editors, *Studia Orientalia Electronica*, volume 113, pages 197–208. Finnish Oriental Society, WS Bookwell Oy, Helsinki, Finland.

Jörg Tiedemann. 2012. Parallel data, tools and interfaces in OPUS. In *Proceedings of the Eighth International Conference on Language Resources and Evaluation (LREC'12)*, pages 2214–2218, Istanbul, Turkey. European Language Resources Association (ELRA).

Zubair Torwali. 2019. Early writing in Torwali in Pakistan. In Ari Sherris and Joy Kreeft Peyton, editors, *Teaching Writing to Children in Indigenous Languages: Instructional Practices from Global Contexts*, pages 44–70. Routledge, New York.

Naeem Uddin and Jalal Uddin. 2019. A step towards Torwali machine translation: an analysis of morphosyntactic challenges in a low-resource language. In *Proceedings of the 2nd Workshop on Technologies for MT of Low Resource Languages*, pages 6–10, Dublin, Ireland. European Association for Machine Translation.

Unicode Consortium. 2021. Arabic. In *The Unicode Standard (Version 14.0.0)*, chapter 9.2, pages 373–398. Unicode Consortium, Mountain View, CA.

Kees Versteegh. 2001. Arabic in Madagascar. *Bulletin of the School of Oriental and African Studies*, 64(2):177–187.

Bernard L. Welch. 1947. The generalization of "Student's" problem when several different population varlances are involved. *Biometrika*, 34(1/2):28–35.

Guillaume Wenzek, Vishrav Chaudhary, Angela Fan, Sahir Gomez, Naman Goyal, Somya Jain, Douwe Kiela, Tristan Thrush, and Francisco Guzmán. 2021. Findings of the WMT 2021 shared task on large-scale multilingual machine translation. In *Proceedings of the Sixth Conference on Machine Translation*, pages 89–99, Online. Association for Computational Linguistics.

Ken Whistler. 2021. Unicode normalization forms. Technical Report TR15-51, Unicode Consortium. Version 14.0.0.

Jens Wilkens. 2016. Buddhism in the West Uyghur kingdom and beyond. In Carmen Meinert, editor, *Transfer of Buddhism across Central Asian networks (7th to 13th centuries)*, volume 8 of *Dynamics in the History of Religions*, chapter 6, pages 189–249. Brill.

André Wink. 1991. *Al-Hind: The Making of the Indo-Islamic World: Early Medieval India and the Expansion of Islam 7th–11th Centuries*, volume 1. Brill, Leiden & New York.

Richard Olaf Winstedt. 1961. Malay chronicles from Sumatra and Malaya. In D. G. E. Hall, editor, *Historians of South-East Asia*, volume II of *Historical Writing on the Peoples of Asia*, pages 24–28. Oxford University Press, Oxford, UK.

Linting Xue, Aditya Barua, Noah Constant, Rami Al-Rfou, Sharan Narang, Mihir Kale, Adam Roberts, and Colin Raffel. 2022. ByT5: Towards a token-free future with pre-trained byte-to-byte models. *Transactions of the Association for Computational Linguistics*, 10:291–306.

Linting Xue, Noah Constant, Adam Roberts, Mihir Kale, Rami Al-Rfou, Aditya Siddhant, Aditya Barua, and Colin Raffel. 2021. mT5: A massively multilingual pre-trained text-to-text transformer. In *Proceedings of the 2021 Conference of the North American Chapter of the Association for Computational Linguistics: Human Language Technologies*, pages 483–498, Online. Association for Computational Linguistics.

Mahire Yakup, Wayit Abliz, Joan Sereno, and Manuel Perea. 2015. Extending models of visual-word recognition to semicursive scripts: Evidence from masked priming in Uyghur. *Journal of Experimental Psychology: Human Perception and Performance*, 41(6):1553–1562.

Ehsan Yarshater. 2011. The Iranian language of Azerbaijan. In Ehsan Yarshater, editor, *Encyclopædia Iranica*, volume III/3, pages 238–245. Brill, Leiden, The Netherlands. Online.

Rana Yassin, David L. Share, and Yasmin Shalhoub-Awwad. 2020. Learning to spell in Arabic: The impact of script-specific visual-orthographic features. *Frontiers in Psychology*, 11:2059.

Altaf Hussain Yatoo. 2012. *The Islamization of Kashmir (A Study of Muslim Missionaries)*. Gulshan Books, Srinagar, Jammu and Kashmir, India.

Sandy L. Zabell. 2008. On Student's 1908 article "The Probable Error of a Mean". *Journal of the American Statistical Association*, 103(481):1–7.

Biao Zhang, Philip Williams, Ivan Titov, and Rico Sennrich. 2020. Improving massively multilingual neural machine translation and zero-shot translation. In *Proceedings of the 58th Annual Meeting of the Association for Computational Linguistics*, pages 1628–1639, Online. Association for Computational Linguistics.

# A Letter Inventories for Individual Languages

The letter repertoire for the eight languages investigated in this paper — South Azerbaijani (azb), Sorani Kurdish (ckb), Kashmiri (ks), Malay (ms), Western Punjabi (pnb), Sindhi (sd), Uyghur (ug) and Urdu (ur) — is shown in Table 22 below. Overall we identified 118 characters including letters and various diacritics. The table shows, for each character its corresponding Unicode code point, Unicode name and the languages that use it (indicated by the checkmark).

| Character | Code Point | Character Name | Language Tags | | | | | | | |
|---|---|---|---|---|---|---|---|---|---|---|
| | | | azb | ckb | ks | ms | pnb | sd | ug | ur |
| ؠ | U+0620 | Kashmiri Yeh | | | ✓ | | | | | |
| ء | U+0621 | Hamza | | | | ✓ | ✓ | ✓ | | ✓ |
| آ | U+0622 | Alef with Madda Above | ✓ | | ✓ | ✓ | ✓ | ✓ | | ✓ |
| أ | U+0623 | Alef with Hamza Above | | | ✓ | ✓ | ✓ | ✓ | | ✓ |
| ؤ | U+0624 | Waw with Hamza Above | ✓ | | ✓ | ✓ | ✓ | ✓ | | ✓ |
| إ | U+0625 | Alef with Hamza Below | | | ✓ | ✓ | ✓ | | | ✓ |
| ئ | U+0626 | Yeh with Hamza Above | ✓ | ✓ | | | ✓ | ✓ | ✓ | ✓ |
| ا | U+0627 | Alef | ✓ | ✓ | ✓ | ✓ | ✓ | ✓ | ✓ | ✓ |
| ب | U+0628 | Beh | ✓ | ✓ | ✓ | ✓ | ✓ | ✓ | ✓ | ✓ |
| ة | U+0629 | Teh Marbuta | | | | ✓ | ✓ | | ✓ | ✓ |
| ت | U+062A | Teh | ✓ | ✓ | ✓ | ✓ | ✓ | ✓ | ✓ | ✓ |
| ث | U+062B | Theh | ✓ | | ✓ | ✓ | ✓ | ✓ | | ✓ |
| ج | U+062C | Jeem | ✓ | ✓ | ✓ | ✓ | ✓ | ✓ | ✓ | ✓ |
| ح | U+062D | Hah | ✓ | ✓ | ✓ | ✓ | ✓ | ✓ | | ✓ |
| خ | U+062E | Khah | ✓ | ✓ | ✓ | ✓ | ✓ | ✓ | ✓ | ✓ |
| د | U+062F | Dal | ✓ | ✓ | ✓ | ✓ | ✓ | ✓ | ✓ | ✓ |
| ذ | U+0630 | Thal | ✓ | | ✓ | ✓ | ✓ | ✓ | ✓ | ✓ |
| ر | U+0631 | Reh | ✓ | ✓ | ✓ | ✓ | ✓ | ✓ | ✓ | ✓ |
| ز | U+0632 | Zain | ✓ | ✓ | ✓ | ✓ | ✓ | ✓ | ✓ | ✓ |
| س | U+0633 | Seen | ✓ | ✓ | ✓ | ✓ | ✓ | ✓ | ✓ | ✓ |
| ش | U+0634 | Sheen | ✓ | ✓ | ✓ | ✓ | ✓ | ✓ | ✓ | ✓ |
| ص | U+0635 | Sad | ✓ | | ✓ | ✓ | ✓ | ✓ | | ✓ |
| ض | U+0636 | Dad | ✓ | | ✓ | ✓ | ✓ | ✓ | | ✓ |
| ط | U+0637 | Tah | ✓ | | ✓ | ✓ | ✓ | ✓ | | ✓ |
| ظ | U+0638 | Zah | ✓ | | ✓ | ✓ | ✓ | ✓ | | ✓ |
| ع | U+0639 | Ain | | ✓ | ✓ | ✓ | ✓ | ✓ | ✓ | ✓ |
| غ | U+063A | Ghain | ✓ | ✓ | ✓ | ✓ | ✓ | ✓ | ✓ | ✓ |
| ؽ | U+063D | Farsi Yeh with Inverted V | ✓ | | | | | | | |
| ف | U+0641 | Feh | ✓ | ✓ | ✓ | ✓ | ✓ | ✓ | ✓ | ✓ |
| ق | U+0642 | Qaf | ✓ | ✓ | ✓ | ✓ | ✓ | ✓ | ✓ | ✓ |
| ك | U+0643 | Kaf | | | | | | | ✓ | ✓ |
| ل | U+0644 | Lam | ✓ | ✓ | ✓ | ✓ | ✓ | ✓ | ✓ | ✓ |
| م | U+0645 | Meem | ✓ | ✓ | ✓ | ✓ | ✓ | ✓ | ✓ | ✓ |
| ن | U+0646 | Noon | ✓ | ✓ | ✓ | ✓ | ✓ | ✓ | ✓ | ✓ |
| ه | U+0647 | Heh | ✓ | ✓ | | ✓ | ✓ | | | ✓ |
| و | U+0648 | Waw | ✓ | ✓ | ✓ | ✓ | ✓ | ✓ | ✓ | ✓ |
| ى | U+0649 | Alef Maksura | | ✓ | | ✓ | ✓ | | ✓ | ✓ |
| ي | U+064A | Yeh | | | | ✓ | | ✓ | ✓ | |
| ً | U+064B | Fathatan | | | | | ✓ | | | ✓ |
| ٌ | U+064C | Dammatan | | | | | ✓ | | | ✓ |
| ٍ | U+064D | Kasratan | | | | | ✓ | | | ✓ |

*Continued on next page*

Table 22 – *Continued from previous page*

| Character | Code Point | Character Name | az | ckb | ks | ms | pa | sd | ug | ur |
|---|---|---|---|---|---|---|---|---|---|---|
| | | | Language Tags | | | | | | | |
| َ | U+064E | Fatha | ✓ | | ✓ | ✓ | ✓ | ✓ | ✓ | ✓ |
| ُ | U+064F | Damma | ✓ | | ✓ | ✓ | ✓ | ✓ | ✓ | ✓ |
| ِ | U+0650 | Kasra | ✓ | | ✓ | ✓ | ✓ | ✓ | ✓ | ✓ |
| ّ | U+0651 | Shadda | ✓ | | ✓ | ✓ | ✓ | ✓ | | ✓ |
| ْ | U+0652 | Sukun | ✓ | | ✓ | ✓ | ✓ | | | ✓ |
| ٓ | U+0653 | Maddah Above | ✓ | | ✓ | ✓ | ✓ | ✓ | | ✓ |
| ٔ | U+0654 | Hamza Above | | | ✓ | | ✓ | ✓ | ✓ | ✓ |
| ٕ | U+0655 | Hamza Below | | | ✓ | | ✓ | ✓ | | ✓ |
| ٖ | U+0656 | Subscript Alef | | | ✓ | | ✓ | | | |
| ٗ | U+0657 | Inverted Damma | | | ✓ | | ✓ | | | ✓ |
| ٚ | U+065A | Vowel Sign Small V Above | | | ✓ | | | | | |
| ٟ | U+065F | Wavy Hamza Below | | | ✓ | | | | | |
| ٰ | U+0670 | Superscript Alef | | | | | ✓ | | | ✓ |
| ٱ | U+0671 | Alef Wasla | ✓ | | | | | | | |
| ٲ | U+0672 | Alef with Wavy Hamza Above | | | ✓ | | | | | |
| ٳ | U+0673 | Alef with Wavy Hamza Below | | | ✓ | | | | | |
| ٹ | U+0679 | Tteh | | | ✓ | | ✓ | | | ✓ |
| ٺ | U+067A | Tteheh | | | | | | ✓ | | |
| ٻ | U+067B | Beeh | | | | | | ✓ | | |
| ٽ | U+067D | Teh with Three Dots Above Downwards | | | | | | ✓ | | |
| پ | U+067E | Peh | ✓ | ✓ | ✓ | | ✓ | ✓ | ✓ | ✓ |
| ٿ | U+067F | Teheh | | | | | | ✓ | | |
| ڀ | U+0680 | Beheh | | | | | | ✓ | | |
| ڃ | U+0683 | Nyeh | | | | | | ✓ | | |
| ڄ | U+0684 | Dyeh | | | | | | ✓ | | |
| څ | U+0685 | Hah with Three Dots Above | | | | | | | ✓ | |
| چ | U+0686 | Tcheh | ✓ | ✓ | ✓ | ✓ | ✓ | ✓ | ✓ | ✓ |
| ڇ | U+0687 | Tcheheh | | | | | | ✓ | | |
| ڈ | U+0688 | Ddal | | | ✓ | | ✓ | | | ✓ |
| ڊ | U+068A | Dal with Dot Below | | | | | | ✓ | | |
| ڌ | U+068C | Dahal | | | | | | ✓ | | |
| ڍ | U+068D | Ddahal | | | | | | ✓ | | |
| ڏ | U+068F | Dal with Three Dots Above Downwards | | | | | | ✓ | | |
| ڑ | U+0691 | Rreh | | | ✓ | | ✓ | | | ✓ |
| ڕ | U+0695 | Reh with Small V Below | | ✓ | | | | | | |
| ژ | U+0698 | Jeh | ✓ | ✓ | ✓ | | ✓ | | ✓ | ✓ |
| ڙ | U+0699 | Reh with Four Dots Above | | | | | | ✓ | | |
| ڠ | U+06A0 | Ain with Three Dots Above | | | | ✓ | | | | |
| ڤ | U+06A4 | Veh | | ✓ | | ✓ | | | | |
| ڦ | U+06A6 | Peheh | | | | | | ✓ | | |
| ک | U+06A9 | Keheh | ✓ | ✓ | ✓ | ✓ | ✓ | ✓ | | ✓ |
| ڪ | U+06AA | Swash Kaf | | | | | | ✓ | | |
| ڭ | U+06AD | Ng | | | | | | | ✓ | |
| گ | U+06AF | Gaf | ✓ | ✓ | ✓ | | ✓ | ✓ | ✓ | ✓ |
| ڱ | U+06B1 | Ngoeh | | | | | | ✓ | | |
| ڳ | U+06B3 | Gueh | | | | | | ✓ | | |
| ڴ | U+06B4 | Gaf with Three Dots Above | ✓ | | | | | | | |
| ڵ | U+06B5 | Lam with Small V | | ✓ | | | | | | |
| ں | U+06BA | Noon Ghunna | | | ✓ | | ✓ | | | ✓ |
| ڻ | U+06BB | Rnoon | | | | | | ✓ | | |
| ڽ | U+06BD | Noon with Three Dots Above | | | | ✓ | | | | |
| ھ | U+06BE | Heh Doachashmee | | ✓ | ✓ | | ✓ | ✓ | ✓ | ✓ |
| ۀ | U+06C0 | Heh with Yeh Above | | ✓ | | | | | | |
| ہ | U+06C1 | Heh Goal | | | ✓ | | ✓ | | | ✓ |
| ۂ | U+06C2 | Heh Goal with Hamza Above | | | ✓ | | ✓ | | | ✓ |
| ۃ | U+06C3 | Teh Marbuta Goal | | | | | ✓ | | | ✓ |
| ۄ | U+06C4 | Waw with Ring | | | ✓ | | | | | |
| ۆ | U+06C6 | Oe | ✓ | ✓ | ✓ | | | | ✓ | |
| ۇ | U+06C7 | U | ✓ | | | | | | ✓ | |
| ۈ | U+06C8 | Yu | | | | | | | ✓ | |
| ۊ | U+06CA | Waw with Two Dots Above | | ✓ | | | | | | |

*Continued on next page*

Table 22 – *Continued from previous page*

| Character | Code Point | Character Name | Language Tags az | ckb | ks | ms | pa | sd | ug | ur |
|---|---|---|---|---|---|---|---|---|---|---|
| ۋ | U+06CB | Ve | | | | | | | ✓ | |
| ی | U+06CC | Farsi Yeh | ✓ | ✓ | ✓ | | ✓ | | | ✓ |
| ێ | U+06CE | Yeh with Small V | ✓ | ✓ | | | | | | |
| ۏ | U+06CF | Waw with Dot Above | | | | ✓ | | | | |
| ې | U+06D0 | E | | | | | | | ✓ | |
| ے | U+06D2 | Yeh Barree | | | | | ✓ | | | ✓ |
| ۓ | U+06D3 | Yeh Barree with Hamza Above | | | | | ✓ | | | ✓ |
| ە | U+06D5 | Ae | | ✓ | | | | | ✓ | |
| ۽ | U+06FD | Sign Sindhi Ampersand | | | | | | ✓ | | |
| ۾ | U+06FE | Sign Sindhi Postposition Men | | | | | | ✓ | | |
| ݢ | U+0762 | Keheh with Dot Above | | | | ✓ | | | | |
| ݢ | U+0762 | Keheh with Dot Above | | | | ✓ | | | | |
| ݣ | U+0763 | Keheh with Three Dots Above | | | | ✓ | | | | |
| ݨ | U+0768 | Noon with Small Tah Above | | | | | ✓ | | | |
| ݬ | U+076C | Reh with Hamza Above | | | ✓ | | | | | |
| ࣇ | U+08C7 | Lam with Small Tah Above | | | | | ✓ | | | |

Table 22: Letter inventories for individual languages.

# B Language model experiments

## B.1 Character language models

Full character language model results are shown here for Kashmiri (Table 23), Kurdish Sorani (Table 24), Malay (Table 25), Punjabi Shahmukhi (Table 26), Sindhi (Table 27), South Azerbaijani (Table 28), Uyghur (Table 29), and Urdu (Table 30).

Assuming the significance level of $\alpha = 0.05$, the results for Kashmiri in Table 23, which is the smallest dataset, indicate that cross-entropy improvements for $n$-gram orders 4, 9, 10 are statistically significant. For the 8-grams, the MW and BW tests indicate borderline significance disagreeing with the WS test. The discrepancy observed between lower orders is probably due to overfitting as the dataset is tiny. Punjabi (Shahmukhi) and Kurdish (Sorani) are the second and third biggest datasets, respectively, and the results in Table 26 and Table 24 indicate statistically significant improvements across the board, with all the three tests agreeing with each other. Sindhi (Table 27) is similar in some respects to Kashmiri in that the low $n$-gram orders of 3 and 4 are not very reliable, while the results for the rest of the orders indicate significant improvements. While overfitting may play a certain role, upon informal inspection it appears that, similar to Kashmiri, the Sindhi dataset is quite noisy, even after filtering. Finally, the results for Urdu in Table 30 indicate that, similar to Kurdish (Sorani), Punjabi (Shahmukhi), South Azerbaijani and Uyghur the improvements are statistically significant across the board. The best results are obtained for Malay (Table 25) with up to 2.9% improvement in character entropy.

## B.2 Word language models

Full word language model results are shown here for Kashmiri (Table 31), Kurdish Sorani (Table 32), Malay (Table 33), Punjabi Shahmukhi (Table 34), Sindhi (Table 35), South Azerbaijani (Table 36), Uyghur (Table 37), and Urdu (Table 38).

As can be seen from Table 31, the hypothesis testing for Kashmiri shows that the null hypothesis is confirmed by all the three algorithms for all the orders $n$ most of the time. This is evident from the tests' $p$-values (these exceed the significance level $\alpha = 0.05$) as well as the $t$-test confidence intervals that contain the null hypothesis, which in our case corresponds to the zero difference in means. This is likely the artifact of the models overfitting a very small dataset, even for relatively less scarce bigrams. It is interesting to note that for the rest of the languages the alternative hypotheses for all the models is uniformly confirmed: the small decrease in cross-entropy expressed as bits per word observed for all the languages and configurations is statistically significant.

| n | Δμ | | WS | | | | MW | | BM | |
|---|---|---|---|---|---|---|---|---|---|---|
| | Δ | % | L | H | t | p | t | p | t | p |
| 3 | −0.003 | 0.08 | −0.015 | 0.009 | −0.458 | 0.647 | 5242.0 | 0.555 | −0.588 | 0.557 |
| 4 | −0.021 | 0.71 | −0.036 | −0.006 | −2.722 | 0.007 | 5946.0 | 0.021 | −2.356 | 0.019 |
| 5 | −0.006 | 0.21 | −0.024 | 0.013 | −0.607 | 0.544 | 5247.0 | 0.547 | −0.601 | 0.549 |
| 6 | −0.006 | 0.22 | −0.023 | 0.012 | −0.652 | 0.515 | 5327.0 | 0.425 | −0.793 | 0.429 |
| 7 | −0.005 | 0.2 | −0.024 | 0.013 | −0.578 | 0.564 | 5276.0 | 0.501 | −0.671 | 0.503 |
| 8 | −0.016 | 0.63 | −0.034 | 0.002 | −1.791 | 0.075 | 5802.0 | 0.05 | −1.977 | 0.05 |
| 9 | −0.022 | 0.85 | −0.04 | −0.004 | −2.468 | 0.014 | 6007.0 | 0.014 | −2.51 | 0.013 |
| 10 | −0.028 | 1.07 | −0.046 | −0.009 | −2.96 | 0.003 | 6065.0 | 0.009 | −2.676 | 0.008 |

Table 23: Significance tests for Kashmiri character *n*-gram language models.

| n | Δμ | | WS | | | | MW | | BM | |
|---|---|---|---|---|---|---|---|---|---|---|
| | Δ | % | L | H | t | p | t | p | t | p |
| 3 | −0.01 | 0.36 | −0.011 | −0.009 | −22.395 | 0.0 | 9916.0 | 0.0 | −115.778 | 0.0 |
| 4 | −0.003 | 0.16 | −0.004 | −0.002 | −6.687 | 0.0 | 7535.0 | 0.0 | −7.394 | 0.0 |
| 5 | −0.004 | 0.24 | −0.005 | −0.003 | −6.288 | 0.0 | 7369.0 | 0.0 | −6.724 | 0.0 |
| 6 | −0.003 | 0.2 | −0.004 | −0.002 | −4.633 | 0.0 | 6718.0 | 0.0 | −4.547 | 0.0 |
| 7 | −0.004 | 0.28 | −0.005 | −0.002 | −5.768 | 0.0 | 7270.0 | 0.0 | −6.294 | 0.0 |
| 8 | −0.006 | 0.44 | −0.007 | −0.004 | −8.596 | 0.0 | 8040.0 | 0.0 | −10.003 | 0.0 |
| 9 | −0.005 | 0.43 | −0.006 | −0.004 | −8.741 | 0.0 | 8006.0 | 0.0 | −9.734 | 0.0 |
| 10 | −0.005 | 0.39 | −0.006 | −0.003 | −7.075 | 0.0 | 7623.0 | 0.0 | −7.86 | 0.0 |

Table 24: Significance tests for Kurdish (Sorani) character *n*-gram language models.

| n | Δμ | | WS | | | | MW | | BM | |
|---|---|---|---|---|---|---|---|---|---|---|
| | Δ | % | L | H | t | p | t | p | t | p |
| 3 | −0.065 | 1.818 | −0.066 | −0.064 | −194.951 | 0.0 | 10 000.0 | 0.0 | −∞ | 0.0 |
| 4 | −0.062 | 2.036 | −0.063 | −0.061 | −143.242 | 0.0 | 10 000.0 | 0.0 | −∞ | 0.0 |
| 5 | −0.06 | 2.232 | −0.061 | −0.059 | −135.144 | 0.0 | 10 000.0 | 0.0 | −∞ | 0.0 |
| 6 | −0.064 | 2.526 | −0.065 | −0.063 | −132.604 | 0.0 | 10 000.0 | 0.0 | −∞ | 0.0 |
| 7 | −0.067 | 2.736 | −0.068 | −0.066 | −135.678 | 0.0 | 10 000.0 | 0.0 | −∞ | 0.0 |
| 8 | −0.07 | 2.885 | −0.071 | −0.069 | −126.796 | 0.0 | 10 000.0 | 0.0 | −∞ | 0.0 |
| 9 | −0.07 | 2.931 | −0.071 | −0.069 | −116.01 | 0.0 | 10 000.0 | 0.0 | −∞ | 0.0 |
| 10 | −0.07 | 2.922 | −0.071 | −0.068 | −103.192 | 0.0 | 10 000.0 | 0.0 | −∞ | 0.0 |

Table 25: Significance tests for Malay character *n*-gram language models.

| n | Δμ | | WS | | | | MW | | BM | |
|---|---|---|---|---|---|---|---|---|---|---|
| | Δ | % | L | H | t | p | t | p | t | p |
| 3 | −0.011 | 0.32 | −0.012 | −0.01 | −34.213 | 0.0 | 10 000.0 | 0.0 | −∞ | 0.0 |
| 4 | −0.013 | 0.44 | −0.013 | −0.012 | −35.959 | 0.0 | 10 000.0 | 0.0 | −∞ | 0.0 |
| 5 | −0.01 | 0.39 | −0.011 | −0.009 | −24.399 | 0.0 | 9941.0 | 0.0 | −167.411 | 0.0 |
| 6 | −0.008 | 0.34 | −0.009 | −0.007 | −16.158 | 0.0 | 9667.0 | 0.0 | −29.955 | 0.0 |
| 7 | −0.007 | 0.33 | −0.008 | −0.006 | −13.856 | 0.0 | 9429.0 | 0.0 | −21.268 | 0.0 |
| 8 | −0.006 | 0.3 | −0.008 | −0.004 | −6.491 | 0.0 | 8928.0 | 0.0 | −14.653 | 0.0 |
| 9 | −0.007 | 0.36 | −0.009 | −0.006 | −7.799 | 0.0 | 9061.0 | 0.0 | −16.395 | 0.0 |
| 10 | −0.007 | 0.34 | −0.009 | −0.005 | −6.11 | 0.0 | 8609.0 | 0.0 | −12.104 | 0.0 |

Table 26: Significance tests for Punjabi (Shahmukhi) character *n*-gram language models.

| $n$ | $\Delta_\mu$ | | WS | | | | MW | | BM | |
|---|---|---|---|---|---|---|---|---|---|---|
| | $\Delta$ | % | $L$ | $H$ | $t$ | $p$ | $t$ | $p$ | $t$ | $p$ |
| 3 | 0.001 | −0.03 | −0.002 | 0.004 | 0.71 | 0.479 | 4649.0 | 0.392 | 0.853 | 0.395 |
| 4 | 0.011 | −0.28 | 0.005 | 0.016 | 3.628 | 0.0 | 3461.0 | 0.0 | 3.935 | 0.0 |
| 5 | −0.018 | 0.47 | −0.028 | −0.008 | −3.451 | 0.001 | 6718.0 | 0.0 | −4.51 | 0.0 |
| 6 | −0.028 | 0.79 | −0.039 | −0.018 | −5.183 | 0.0 | 7477.0 | 0.0 | −7.109 | 0.0 |
| 7 | −0.016 | 0.46 | −0.025 | −0.006 | −3.309 | 0.001 | 6714.0 | 0.0 | −4.448 | 0.0 |
| 8 | −0.024 | 0.71 | −0.033 | −0.014 | −4.775 | 0.0 | 6864.0 | 0.0 | −4.977 | 0.0 |
| 9 | −0.015 | 0.45 | −0.026 | −0.004 | −2.715 | 0.007 | 6119.0 | 0.006 | −2.814 | 0.005 |
| 10 | −0.014 | 0.42 | −0.026 | −0.003 | −2.443 | 0.015 | 5889.0 | 0.03 | −2.212 | 0.028 |

Table 27: Significance tests for Sindhi character $n$-gram language models.

| $n$ | $\Delta_\mu$ | | WS | | | | MW | | BM | |
|---|---|---|---|---|---|---|---|---|---|---|
| | $\Delta$ | % | $L$ | $H$ | $t$ | $p$ | $t$ | $p$ | $t$ | $p$ |
| 3 | −0.012 | 0.418 | −0.013 | −0.011 | −24.371 | 0.0 | 9953.0 | 0.0 | −192.439 | 0.0 |
| 4 | −0.004 | 0.186 | −0.005 | −0.003 | −6.649 | 0.0 | 7482.0 | 0.0 | −7.222 | 0.0 |
| 5 | −0.002 | 0.131 | −0.003 | −0.001 | −3.133 | 0.002 | 6137.0 | 0.005 | −2.866 | 0.005 |
| 6 | −0.001 | 0.093 | −0.003 | 0.0 | −1.993 | 0.048 | 5837.0 | 0.041 | −2.072 | 0.04 |
| 7 | −0.002 | 0.14 | −0.003 | 0.0 | −2.526 | 0.012 | 5895.0 | 0.029 | −2.227 | 0.027 |
| 8 | −0.002 | 0.143 | −0.003 | 0.0 | −2.52 | 0.013 | 5997.0 | 0.015 | −2.489 | 0.014 |
| 9 | −0.002 | 0.13 | −0.003 | 0.0 | −2.348 | 0.02 | 5945.0 | 0.021 | −2.352 | 0.02 |
| 10 | −0.002 | 0.171 | −0.004 | −0.001 | −2.889 | 0.004 | 6238.0 | 0.002 | −3.107 | 0.002 |

Table 28: Significance tests for South Azerbaijani character $n$-gram language models.

| $n$ | $\Delta_\mu$ | | WS | | | | MW | | BM | |
|---|---|---|---|---|---|---|---|---|---|---|
| | $\Delta$ | % | $L$ | $H$ | $t$ | $p$ | $t$ | $p$ | $t$ | $p$ |
| 3 | −0.002 | 0.074 | −0.003 | −0.001 | −4.42 | 0.0 | 6636.0 | 0.0 | −4.273 | 0.0 |
| 4 | −0.001 | 0.051 | −0.002 | 0.0 | −2.257 | 0.025 | 5801.0 | 0.05 | −1.981 | 0.049 |
| 5 | −0.004 | 0.203 | −0.005 | −0.003 | −7.131 | 0.0 | 7690.0 | 0.0 | −7.964 | 0.0 |
| 6 | −0.004 | 0.219 | −0.005 | −0.003 | −6.315 | 0.0 | 7394.0 | 0.0 | −6.829 | 0.0 |
| 7 | −0.003 | 0.164 | −0.004 | −0.002 | −4.348 | 0.0 | 6661.0 | 0.0 | −4.353 | 0.0 |
| 8 | −0.004 | 0.225 | −0.005 | −0.002 | −5.126 | 0.0 | 6907.0 | 0.0 | −5.133 | 0.0 |
| 9 | −0.004 | 0.24 | −0.006 | −0.002 | −4.549 | 0.0 | 6699.0 | 0.0 | −4.476 | 0.0 |
| 10 | −0.005 | 0.295 | −0.006 | −0.003 | −6.446 | 0.0 | 7460.0 | 0.0 | −6.977 | 0.0 |

Table 29: Significance tests for Uyghur character $n$-gram language models.

| $n$ | $\Delta_\mu$ | | WS | | | | MW | | BM | |
|---|---|---|---|---|---|---|---|---|---|---|
| | $\Delta$ | % | $L$ | $H$ | $t$ | $p$ | $t$ | $p$ | $t$ | $p$ |
| 3 | −0.003 | 0.11 | −0.004 | −0.001 | −4.372 | 0.0 | 6646.0 | 0.0 | −4.306 | 0.0 |
| 4 | −0.005 | 0.22 | −0.005 | −0.004 | −17.256 | 0.0 | 9575.0 | 0.0 | −34.762 | 0.0 |
| 5 | −0.004 | 0.2 | −0.004 | −0.003 | −9.06 | 0.0 | 8813.0 | 0.0 | −15.737 | 0.0 |
| 6 | −0.004 | 0.26 | −0.006 | −0.003 | −5.352 | 0.0 | 8363.0 | 0.0 | −11.571 | 0.0 |
| 7 | −0.004 | 0.24 | −0.005 | −0.002 | −4.601 | 0.0 | 8246.0 | 0.0 | −10.74 | 0.0 |
| 8 | −0.002 | 0.14 | −0.004 | 0.0 | −2.35 | 0.02 | 8220.0 | 0.0 | −10.431 | 0.0 |
| 9 | −0.002 | 0.13 | −0.005 | 0.001 | −1.24 | 0.217 | 7896.0 | 0.0 | −8.207 | 0.0 |
| 10 | −0.005 | 0.37 | −0.008 | −0.002 | −3.422 | 0.001 | 8344.0 | 0.0 | −10.745 | 0.0 |

Table 30: Significance tests for Urdu character $n$-gram language models.

| $n$ | $\Delta_\mu$ | | WS | | | | MW | | BM | |
|---|---|---|---|---|---|---|---|---|---|---|
| | $\Delta$ | % | $L$ | $H$ | $t$ | $p$ | $t$ | $p$ | $t$ | $p$ |
| 2 | 0.034 | −0.33 | −0.008 | 0.076 | 1.595 | 0.112 | 4301.0 | 0.088 | 1.72 | 0.087 |
| 3 | −0.039 | 0.39 | −0.088 | 0.011 | −1.531 | 0.127 | 5589.0 | 0.15 | −1.447 | 0.149 |
| 4 | −0.022 | 0.22 | −0.067 | 0.023 | −0.957 | 0.34 | 5308.0 | 0.452 | −0.751 | 0.454 |
| 5 | 0.017 | −0.17 | −0.029 | 0.063 | 0.727 | 0.468 | 4817.0 | 0.656 | 0.443 | 0.659 |

Table 31: Significance tests for Kashmiri word $n$-gram language models.

| $n$ | $\Delta_\mu$ | | WS | | | | MW | | BM | |
|---|---|---|---|---|---|---|---|---|---|---|
| | $\Delta$ | % | $L$ | $H$ | $t$ | $p$ | $t$ | $p$ | $t$ | $p$ |
| 2 | −0.031 | 0.41 | −0.038 | −0.024 | −8.678 | 0.0 | 8058.0 | 0.0 | −10.074 | 0.0 |
| 3 | −0.034 | 0.49 | −0.04 | −0.028 | −10.393 | 0.0 | 8492.0 | 0.0 | −13.09 | 0.0 |
| 4 | −0.034 | 0.49 | −0.041 | −0.027 | −9.297 | 0.0 | 8237.0 | 0.0 | −11.275 | 0.0 |
| 5 | −0.035 | 0.5 | −0.042 | −0.028 | −10.406 | 0.0 | 8464.0 | 0.0 | −13.209 | 0.0 |

Table 32: Significance tests for Kurdish (Sorani) word $n$-gram language models.

While the relative improvements for Kurdish Sorani (Table 32), Punjabi Shahmukhi (Table 34), South Azerabaijani (Table 36), Uyghur (Table 37) and Urdu (Table 38) are relatively tiny, possibly due to the relatively small number of modifications compared to the overall size of the datasets, the relative improvements to Sindhi models (Table 35) are over one percent for all the configurations. This may indeed be correlated with the highest number of per-word token modifications for Sindhi among all the languages, denoted $R_w$ in Table 12. The best results are obtained for Malay (Table 33), with up to 3.5% improvement in word entropy.

| $n$ | $\Delta_\mu$ | | WS | | | | MW | | BM | |
|---|---|---|---|---|---|---|---|---|---|---|
| | $\Delta$ | % | $L$ | $H$ | $t$ | $p$ | $t$ | $p$ | $t$ | $p$ |
| 2 | −0.358 | 2.935 | −0.362 | −0.355 | −185.4 | 0.0 | 10 000.0 | 0.0 | $-\infty$ | 0.0 |
| 3 | −0.394 | 3.319 | −0.399 | −0.389 | −166.053 | 0.0 | 10 000.0 | 0.0 | $-\infty$ | 0.0 |
| 4 | −0.403 | 3.411 | −0.408 | −0.398 | −166.567 | 0.0 | 10 000.0 | 0.0 | $-\infty$ | 0.0 |
| 5 | −0.411 | 3.479 | −0.416 | −0.406 | −170.479 | 0.0 | 10 000.0 | 0.0 | $-\infty$ | 0.0 |

Table 33: Significance tests for Malay word $n$-gram language models.

| $n$ | $\Delta_\mu$ | | WS | | | | MW | | BM | |
|---|---|---|---|---|---|---|---|---|---|---|
| | $\Delta$ | % | $L$ | $H$ | $t$ | $p$ | $t$ | $p$ | $t$ | $p$ |
| 2 | −0.015 | 0.15 | −0.018 | −0.012 | −10.834 | 0.0 | 8633.0 | 0.0 | −14.295 | 0.0 |
| 3 | −0.017 | 0.19 | −0.021 | −0.014 | −9.338 | 0.0 | 8237.0 | 0.0 | −11.232 | 0.0 |
| 4 | −0.02 | 0.23 | −0.025 | −0.016 | −9.472 | 0.0 | 8318.0 | 0.0 | −11.75 | 0.0 |
| 5 | −0.02 | 0.23 | −0.024 | −0.016 | −8.916 | 0.0 | 8137.0 | 0.0 | −10.521 | 0.0 |

Table 34: Significance tests for Punjabi (Shahmukhi) word $n$-gram language models.

| $n$ | $\Delta_\mu$ | | WS | | | | MW | | BM | |
|---|---|---|---|---|---|---|---|---|---|---|
| | $\Delta$ | % | $L$ | $H$ | $t$ | $p$ | $t$ | $p$ | $t$ | $p$ |
| 2 | −0.159 | 1.06 | −0.171 | −0.146 | −24.103 | 0.0 | 9948.0 | 0.0 | −180.99 | 0.0 |
| 3 | −0.169 | 1.21 | −0.186 | −0.152 | −19.861 | 0.0 | 9753.0 | 0.0 | −54.902 | 0.0 |
| 4 | −0.177 | 1.26 | −0.192 | −0.161 | −22.847 | 0.0 | 9889.0 | 0.0 | −88.059 | 0.0 |
| 5 | −0.184 | 1.32 | −0.2 | −0.168 | −22.57 | 0.0 | 9898.0 | 0.0 | −108.596 | 0.0 |

Table 35: Significance tests for Sindhi word $n$-gram language models.

| $n$ | $\Delta_\mu$ | | WS | | | | MW | | BM | |
|---|---|---|---|---|---|---|---|---|---|---|
| | $\Delta$ | % | $L$ | $H$ | $t$ | $p$ | $t$ | $p$ | $t$ | $p$ |
| 2 | −0.031 | 0.374 | −0.038 | −0.024 | −9.016 | 0.0 | 8113.0 | 0.0 | −10.307 | 0.0 |
| 3 | −0.027 | 0.332 | −0.034 | −0.02 | −7.376 | 0.0 | 7655.0 | 0.0 | −7.913 | 0.0 |
| 4 | −0.031 | 0.397 | −0.038 | −0.025 | −9.711 | 0.0 | 8325.0 | 0.0 | −11.755 | 0.0 |
| 5 | −0.028 | 0.358 | −0.035 | −0.021 | −8.108 | 0.0 | 7797.0 | 0.0 | −8.828 | 0.0 |

Table 36: Significance tests for South Azerbaijani word $n$-gram language models.

| $n$ | $\Delta_\mu$ | | WS | | | | MW | | BM | |
|---|---|---|---|---|---|---|---|---|---|---|
| | $\Delta$ | % | $L$ | $H$ | $t$ | $p$ | $t$ | $p$ | $t$ | $p$ |
| 2 | −0.024 | 0.2 | −0.033 | −0.015 | −5.14 | 0.0 | 6961.0 | 0.0 | −5.297 | 0.0 |
| 3 | −0.038 | 0.332 | −0.05 | −0.027 | −6.721 | 0.0 | 7537.0 | 0.0 | −7.397 | 0.0 |
| 4 | −0.034 | 0.294 | −0.045 | −0.022 | −5.689 | 0.0 | 7177.0 | 0.0 | −6.029 | 0.0 |
| 5 | −0.042 | 0.367 | −0.055 | −0.029 | −6.319 | 0.0 | 7381.0 | 0.0 | −6.774 | 0.0 |

Table 37: Significance tests for Uyghur word $n$-gram language models.

| $n$ | $\Delta_\mu$ | | WS | | | | MW | | BM | |
|---|---|---|---|---|---|---|---|---|---|---|
| | $\Delta$ | % | $L$ | $H$ | $t$ | $p$ | $t$ | $p$ | $t$ | $p$ |
| 2 | −0.006 | 0.08 | −0.008 | −0.003 | −4.923 | 0.0 | 6796.0 | 0.0 | −4.804 | 0.0 |
| 3 | −0.004 | 0.06 | −0.007 | −0.002 | −3.507 | 0.001 | 6394.0 | 0.001 | −3.531 | 0.001 |
| 4 | −0.006 | 0.1 | −0.009 | −0.003 | −4.2 | 0.0 | 6637.0 | 0.0 | −4.285 | 0.0 |
| 5 | −0.007 | 0.1 | −0.009 | −0.004 | −4.927 | 0.0 | 6842.0 | 0.0 | −4.914 | 0.0 |

Table 38: Significance tests for Urdu word $n$-gram language models.